%% file: main.tex
\documentclass[11pt]{article}
\input{preamble}

\input{sections/title}

\begin{document}

% ---------------------------------------------------------------- title and authors
\begin{center}
  {\fontsize{12}{15}\selectfont\bfseries \MakeUppercase{\papertitle}\par}
  \vspace{10pt}

  Christoph Weisser\textsuperscript{a*}\\
  \orcid{0000-0003-0616-1027}\\[3pt]
  Fabian Lukassen\textsuperscript{b,a}\\
  \orcid{0009-0003-5073-5574}
\end{center}

\vspace{4pt}
{\small
\textsuperscript{a}: Hochschule Bielefeld (HSBI) -- University of Applied Sciences and Arts, Bielefeld, Germany.\\
\textsuperscript{b}: Centre for Statistics, Georg-August-Universität Göttingen, Göttingen, Germany.\\[4pt]
\textbf{Correspondence:} christoph.weisser@hsbi.de\par}

% ---------------------------------------------------------------- abstract
\textbf{Abstract}\par
\vspace{-4pt}
\input{sections/abstract}

\textbf{Keywords:} \input{sections/keywords}

% ---------------------------------------------------------------- main text
\input{sections/01_introduction}
\input{sections/02_related_work}
\input{sections/03_method}
\input{sections/04_results}
\input{sections/05_discussion}

% ---------------------------------------------------------------- back matter
\printbibliography[title={References}]

\clearpage
\appendix
\titleformat{\section}{\normalfont\fontsize{12}{14}\selectfont\bfseries}{Appendix \thesection}{0.5em}{}
\input{sections/appendix}

\end{document}

%% file: preamble.tex
\usepackage[a4paper, top=2cm, bottom=1.5cm, left=1.5cm, right=1.5cm]{geometry}
\usepackage[T1]{fontenc}
\usepackage[utf8]{inputenc}
\usepackage{newtxtext, newtxmath}       % Times text and maths
\usepackage{microtype}
\usepackage[english]{babel}
\usepackage{csquotes}

\usepackage{amsmath}
\usepackage{booktabs}
\usepackage{tabularx}
\usepackage{multirow}
\usepackage{array}
\usepackage{graphicx}
\usepackage{xcolor}
\usepackage{enumitem}
\usepackage{titlesec}
\usepackage[labelfont=bf, textfont=it, labelsep=newline, justification=raggedright,
            singlelinecheck=false, skip=3pt]{caption}
\usepackage[style=apa, backend=biber, uniquename=false]{biblatex}

\usepackage[colorlinks=true, linkcolor=black, citecolor=black, urlcolor=blue]{hyperref}
\usepackage{xurl}                          % allow line breaks anywhere in URLs

\setlist{itemsep=2pt, topsep=2pt, parsep=0pt}

\titleformat{\section}{\normalfont\fontsize{12}{14}\selectfont\bfseries}{\thesection}{0.5em}{}
\titleformat{\subsection}{\normalfont\normalsize\bfseries}{\thesubsection}{0.5em}{}
\titleformat{\subsubsection}[runin]{\normalfont\normalsize\bfseries}{}{0pt}{}[.]
\titlespacing*{\section}{0pt}{8pt}{2pt}
\titlespacing*{\subsection}{0pt}{6pt}{0pt}
\titlespacing*{\subsubsection}{0pt}{6pt}{0.5em}

\usepackage{threeparttable}
\usepackage{longtable}
\usepackage{float}
\usepackage{etoolbox}

\newcommand{\tabnote}[1]{\par\vspace{3pt}{\footnotesize\raggedright\textit{Note.} #1\par}}

\newcommand{\todourl}[1]{\textcolor{red}{\url{#1}}}

\usepackage{listings}
\usepackage{tcolorbox}
\tcbuselibrary{breakable,skins,listings}
\definecolor{cpInk}{HTML}{1F2328}
\definecolor{cpRule}{HTML}{9AA0A6}
\definecolor{cpWash}{HTML}{F4F5F7}
\definecolor{cpBlue}{HTML}{2B5C8A}
\definecolor{cpGreen}{HTML}{2E6B4F}
\tcbset{cpbox/.style={enhanced, breakable, boxrule=0.5pt, arc=1.5pt,
  left=5pt, right=5pt, top=6pt, bottom=4pt, fonttitle=\small\bfseries, coltitle=cpInk,
  attach boxed title to top left={yshift=-2mm, xshift=4mm}, boxed title style={boxrule=0.3pt, arc=1pt}}}
\lstdefinestyle{cpprompttext}{basicstyle=\footnotesize\ttfamily\color{cpInk}, breaklines=true,
  columns=fullflexible, keepspaces=true, upquote=true, showstringspaces=false,
  postbreak=\mbox{\textcolor{cpRule}{$\hookrightarrow$}\space}}
\tcbset{cpverb/.style={cpbox, listing only, listing style=cpprompttext}}
\newtcblisting{sysprompt}[1][System prompt]{cpverb, colback=cpWash, colframe=cpRule,
  boxed title style={colback=cpWash!60!white, colframe=cpRule}, title={#1}}
\newtcblisting{userprompt}[1][User message]{cpverb, colback=cpBlue!4!white, colframe=cpBlue!55!white,
  boxed title style={colback=cpBlue!12!white, colframe=cpBlue!55!white}, title={#1}}
\newtcblisting{modelout}[1][Model output]{cpverb, colback=cpGreen!4!white, colframe=cpGreen!55!white,
  boxed title style={colback=cpGreen!12!white, colframe=cpGreen!55!white}, title={#1}}

\newcommand{\orcid}[1]{\href{https://orcid.org/#1}{\small https://orcid.org/#1}}

%% file: sections/title.tex
\newcommand{\papertitle}{Detecting Churn Intent in Customer Chatbots: Typed Decision Models Compared with Language Models and Trained Classifiers}

%% file: sections/abstract.tex
Customers often tell a company's chatbot that they intend to leave. To act on such a message, the
chatbot has to recognise the intention and decide whether to respond automatically or to escalate
the conversation to a member of staff. This needs a correct label, a confidence score that can be
trusted, and an answer that is fast and cheap. Typed decision models are a new class of models for
such decisions. They answer a fixed question about a text with a probability for each answer, and
their vendors advertise low latency, low prices and calibrated probabilities. We evaluate the
flagship typed decision model of every vendor on OpenRouter that accepted our requests, 16 models in
total, against seven prompted language models and two classifiers trained on labelled messages. The
data are English chatbot messages from a public churn corpus. In this corpus, the same message can
signal churn for one provider but not for another: a customer who switches providers is lost to one
and gained by the other. We measure discrimination, calibration, the share of
messages that can be decided automatically at 95\% accuracy, latency and cost, and we analyse the
correctness of every decision with mixed-effects logistic regression. The six best typed decision
models reach F1 scores of 0.93 to 0.95, as high as the best language model, with similar
calibration, and they can decide 96 to 100\% of the messages automatically at 95\% accuracy. Depending on
the model, a typical decision takes 0.26 to 0.60 seconds, and a decision costs at most one fourteenth
of what the best language model charges. Quality varies widely between vendors, however. Some typed
decision models are open language models that were fine-tuned for decisions. On the messages that
signal churn for one provider but not for another, these models tend to be less accurate than the
language models they were built from.

%% file: sections/keywords.tex
Churn intent, typed decision models, large language models, calibration, customer service, chatbots

%% file: sections/01_introduction.tex
\section{Introduction}
\label{sec:introduction}

Churn, the loss of customers who end their contract, directly reduces the revenue of subscription
businesses such as telecom providers. Companies therefore try to recognise customers who are about
to leave, so that they can contact them in time. The quality of these predictions has measurable
financial consequences. In a tournament of churn models built on the same data, differences in
accuracy changed the profit of a retention campaign by hundreds of thousands of dollars \parencite{neslin-2006-defection}. For this reason, churn models are often evaluated by the
profit that they bring \parencite{verbraken-2013-novel}. What customers write is a useful source for such predictions:
e-mails and transcripts of service calls improve churn models beyond contract and usage data
\parencite{decaigny-2020-incorporating,vo-2021-leveraging}.

AI is already used in customer service. In one large study, a generative AI assistant raised the
productivity of support agents by 15\% \parencite{brynjolfsson-2025-generative}. Customers also
write to companies' chatbots, and some of them say directly that they want to leave. A
message such as ``I want to cancel my contract'' or ``I am switching to another provider'' gives the company a chance to keep the customer. It can make a retention offer or escalate the
conversation to a member of staff. To use this chance, the chatbot has to recognise the intention in
a short message and decide what to do with it.

This decision has three requirements. First, the label has to be correct. Second, the system needs a confidence score that can be trusted. A company will act automatically
only on the messages where the model is sure, and it will escalate the others to a member of staff. Third, the decision is
made for every incoming message, so each one has to be fast and cheap.

A company can build such a system in several ways. It can train a classifier on messages that it
has labelled itself, or it can prompt a general-purpose large language model (LLM) to return a label
and a confidence. Since September 2026, it can also use a typed decision model (TDM). A TDM reads a
text together with a question that has a fixed set of answers, and it returns a probability for
each answer instead of generated text \parencite{almeida-2026-introducing,openrouter-2026-decisions}.
The first TDM, TypeSafe's Jev, takes its name from the fast and intuitive ``System~1'' of
dual-process theory \parencite{kahneman-2011-thinking}. Its vendor reports response times of 70 to
500~ms, prices far below those of LLMs, and calibrated probabilities, in the sense that ``higher
confidence means higher accuracy'' \parencite{almeida-2026-introducing}. Within two months, about
17 vendors offered TDMs through OpenRouter, a public model gateway. Several of these models are
open-weight fine-tunes of public LLMs \parencite{openrouter-2026-models}.

These properties fit the three requirements of churn detection, and both TypeSafe and OpenRouter
use churn as an example in their own material. Independent evidence on churn is missing, however.
Recent studies compare TDMs with LLMs and trained classifiers on general intent data, annotation
tasks and synthetic workflows
\parencite{rafe-2026-benchmarking,chen-2026-system1arena,ibrahim-2026-evaluating,roumeliotis-2026-decision},
but none of them evaluates churn.

This paper asks whether companies can use TDMs to detect churn intent in their customer chatbots. We
evaluate the flagship TDM of every vendor on OpenRouter and compare these models with prompted LLMs
from closed and open model families and with two classifiers that are trained on labelled messages.

\textbf{Research question.} \textit{How well do typed decision models detect churn intent in
customer messages, and how do they compare with prompted LLMs and trained classifiers in accuracy,
reliability of confidence, speed and cost?}

\textbf{Contributions.} The paper makes three contributions.
\begin{enumerate}
  \item A comparison of 16 TDMs, seven prompted LLMs and two trained classifiers on churn intent.
    For each system, it reports discrimination, calibration, the share of messages that can be
    decided automatically at a given accuracy, latency and cost per 1,000 decisions. Seven of the TDM versions have not been evaluated in any study that we found.
  \item Two closer analyses with a mixed-effects logistic regression on the correctness of every
    single decision. The first concerns messages that signal churn for one company but not for
    another, for example when a customer announces a switch from one provider to a competitor. The
    second compares open TDMs with the LLMs from which they were fine-tuned, prompted as chat models.
  \item Practical guidance for companies that consider TDMs for churn detection, together with the
    code, prompts and predictions needed to reproduce the results.
\end{enumerate}

Section~\ref{sec:related_work} reviews related work. Section~\ref{sec:method} describes the data,
the systems and the evaluation. Section~\ref{sec:results} reports the results, and
Section~\ref{sec:discussion} discusses what they mean for practice.

%% file: sections/02_related_work.tex
\section{Related Work}
\label{sec:related_work}

\subsection{Churn and Churn Intent in Text}

Churn prediction usually works with behavioural and contractual data. Evaluating churn models by
profit instead of by the area under the ROC curve can change which model is preferred
\parencite{verbraken-2013-novel,manzoor-2025-eprofits}. Text adds information to such models.
Customer e-mails and call transcripts improve churn prediction beyond structured data
\parencite{decaigny-2020-incorporating,vo-2021-leveraging}, and LLMs have been used to extract churn
features from e-mails \parencite{shahabikargar-2025-churnkb}. These studies predict whether a
customer will leave in the future. They do not ask whether a single message states the intention to
leave.

\textcite{amiri-2015-target} introduced the detection of churn intent in short texts, using tweets about three telecom brands in the United States. They treated the task as
target-dependent: the label of a tweet refers to one particular brand, so the same tweet can be churn
for one brand and not for another. They also found that sentiment analysis does not
recognise churn intent \parencite{amiri-2016-short}. \textcite{abbet-2018-churn} extended the task to
chatbot conversations and released the Swisscom corpus that we use. Their neural model was trained
on social media and applied to the chatbot messages without retraining, and it reached an F1 score of
82 to 84 on English. In this corpus, the message ``I want to switch from Swisscom to Salt'' counts as
churn for Swisscom, but not for Salt, which gains the customer. We call such labels
\emph{brand-dependent}: whether a message expresses churn depends on the brand that receives it. \textcite{hadifar-2021-million} showed that churn tweets contain strong
lexical cues, so that a classifier on TF-IDF features is a strong baseline. Most recently,
\textcite{sauter-2026-actionable} classified churn in transcripts of calls to the customer service of a telecom
provider. Zero-shot
GPT-4 reached an F1 score of 0.95 and the provider's fine-tuned production classifier 0.91. Their
test set contains 28 churn cases, and they evaluate neither calibration nor brand-dependent labels.

Brand-dependent labels are an instance of a broader problem. One text can carry different
labels for different targets in target-dependent sentiment analysis \parencite{jiang-2011-target},
aspect-based sentiment analysis \parencite{pontiki-2014-semeval} and stance detection
\parencite{mohammad-2016-semeval}. Systems make more errors when the target of interest is not the target of the
opinion \parencite{mohammad-2016-semeval}. Zero-shot LLMs come close to fine-tuned models on
aspect-based sentiment, but they stay far behind them on stance \parencite{zhang-2024-sentiment}.

\subsection{Prompted Language Models and Trained Classifiers}

Since \textcite{brown-2020-language}, a general LLM can be given a task in its prompt instead of
being trained for it. Whether this is good enough for classification is disputed. ChatGPT labelled
tweets more accurately than crowd workers in four of five annotation tasks
\parencite{gilardi-2023-chatgpt}. In broader comparisons, however, zero-shot LLMs generally do not
reach fine-tuned models \parencite{ziems-2024-can,edwards-2024-language,bucher-2024-finetuned}. The
answer depends on the amount of labelled data. Fine-tuned models need about 100 labelled examples on average to
match prompted LLMs, but on binary tasks they can need thousands to beat zero-shot prompting
\parencite{pecher-2025-comparing}. In a setting with 189 training examples, sentence embeddings with
logistic regression were more accurate than every zero-shot LLM and used a small fraction of their
energy \parencite{zschache-2026-comparing}. For intents in chatbots for customer service,
\textcite{arora-2024-intent} found that the best LLMs were about 8 F1 points better than a classifier
trained on sentence embeddings, which in turn was about 56 times faster. When only the uncertain
cases of the classifier were sent to an LLM, the combination came within about 2\% of the LLM's
accuracy at half of its latency.

The cost and latency of LLMs vary widely. API prices differ by orders of magnitude, and sending easy
queries to cheaper models keeps most of the quality at a fraction of the cost
\parencite{chen-2024-frugalgpt,ong-2025-routellm}. Reasoning models spend many tokens even on easy
inputs \parencite{chen-2025-do}, so the reasoning setting of an LLM affects its cost per decision.
Open models that a company hosts itself can be much cheaper than an API at high utilisation, while
an API can be cheaper at low volume \parencite{irugalbandara-2024-scaling}.

\subsection{Typed Decision Models}

The name ``System One'' refers to dual-process theories. In these theories, fast and intuitive
processes give a default response, and slower reasoning intervenes when the default is in doubt
\parencite{kahneman-2011-thinking,evans-2013-dualprocess}. For LLMs, the terms are used
inconsistently. Surveys call ordinary LLMs System~1 and reasoning models such as DeepSeek-R1
System~2 \parencite{li-2025-from,guo-2025-deepseekr1}. TDM vendors, in contrast, use the term for
models that do not generate text at all.

A TDM receives a state, which is usually a text, together with typed questions. A question can ask
for yes or no, for one of several named options, or for a value on a scale. The model returns a
probability for every admissible answer
\parencite{typesafe-2026-system,openrouter-2026-decisions}. TypeSafe released the first TDM, Jev, in
September 2026 \parencite{almeida-2026-introducing}. Other vendors followed with closed models and
with open-weight models that are fine-tuned from public LLMs such as Qwen
\parencite{chen-2026-introducing,baselabs-2026-mammoth1}.

Most evaluations of TDMs are preprints. \textcite{rafe-2026-benchmarking} compared Jev and seven
open TDMs with open generative LLMs and with trained classifiers on intent detection and other
tasks. A fine-tuned DeBERTa model was the most accurate system when labelled data were available,
and at a guaranteed error rate of 5\%, Jev could decide 53\% of the intent queries automatically.
A larger benchmark found that the largest TDMs are as accurate as fast closed LLMs at lower latency
\parencite{chen-2026-system1arena}. Other studies found that TDMs often overstate their
probabilities \parencite{duan-2026-general}, and that on annotation tasks they are less accurate
than the best LLM but much cheaper \parencite{ibrahim-2026-evaluating}. For the triage of requests to customer service,
\textcite{roumeliotis-2026-decision} found that decision training improved an open model over its
base model on the BANKING77 data, but not on the Bitext data. Several studies describe weaknesses
that matter for churn. TDMs fail most often when a task depends on a labelling rule that the name
of the label does not state \parencite{suppa-2026-do}. The names of the options can override the
definitions that are attached to them \parencite{sun-2026-typesafe}. Jev also has difficulty with the
direction of a stance towards a target \parencite{li-2026-can}.

\subsection{Calibration and Automated Decisions}

A model is calibrated if its probabilities match the frequencies that are observed. Calibration is
assessed with the Brier score \parencite{brier-1950-verification}, with reliability diagrams
\parencite{degroot-1983-comparison,niculescumizil-2005-predicting} and with the expected calibration
error \parencite{naeini-2015-obtaining,guo-2017-calibration}. The expected calibration error depends
on the choice of bins, and it underestimates the error of continuous scores
\parencite{nixon-2019-measuring,kumar-2019-verified}. LLMs can state their confidence as a number.
Such verbalised confidence can be better calibrated than the probabilities of the model's tokens
\parencite{tian-2023-just}, but it clusters at high values and separates correct from incorrect
answers poorly \parencite{xiong-2024-can}. Most of this evidence comes from question answering and
not from classification.

Deciding only the confident cases and passing the rest to a person is the reject option of
\textcite{chow-1970-optimum}, which was later formalised as selective classification
\parencite{elyaniv-2010-foundations,geifman-2017-selective}. Its value depends on how well the
confidence separates correct from incorrect decisions, and this separation can break down when the
data differ from the training data \parencite{kamath-2020-selective}.

In summary, earlier work has compared TDMs with LLMs and trained classifiers several times. None of these studies
evaluates churn or brand-dependent labels, and none covers the current range of vendors. They also compare models with paired bootstrap or McNemar tests. We add a mixed-effects
analysis of the correctness of each decision, as proposed for the evaluation of machine learning
models by \textcite{hagmann-2023-towards}.

%% file: sections/03_method.tex
\section{Method}
\label{sec:method}

The data, the code, the prompts and all predictions are
available at \todourl{https://github.com/fabian-lu/churn-typed-decision-models}.

\subsection{Data}

We use the English chatbot messages from the corpus of \textcite{abbet-2018-churn}, which is
published under a CC BY-SA 3.0 licence at \url{https://github.com/swisscom/churn-intent-DE}.
Participants in a game-like chatbot interface wrote messages that were meant to express churn or
not, and a second person confirmed each label. The messages are therefore short, written by crowd
workers and mostly about Swiss telecom operators. They are not logs of real customer traffic.

We still consider the corpus a suitable test set, for three reasons. First, to our knowledge it is
the only public corpus of chatbot messages with labels for churn intent. Real customer chats are
rarely published, because they contain personal data. Second, the messages resemble what customers
type into a chatbot: they are short, informal and sometimes contain typing errors. Third, every
label was confirmed by a second person, and the corpus contains the difficult case in which a
customer names both the provider that they leave and the one that they join.

Each row of the corpus contains a message, a brand and a label. The label states whether the message
expresses churn from the point of view of that brand. This is the case if the customer intends to
leave, cancel or switch away from the brand, or is seriously considering doing so. Switching to the
brand, complaints without a stated intention to leave, questions and unrelated content are not
churn. For rows without a brand, the label refers to the customer's current provider, which the
message does not name.

Many messages occur in several rows, once for each brand that they are paired with. For most of
these messages, the label is the same in all of their rows; ``I'm not a fan of swisscom'', for
example, is not churn in either of its two rows. For 36 messages, however, the label changes with
the brand. ``I want to switch from Swisscom to Salt'' is churn in the row for Swisscom and not churn
in the row for Salt, because Salt gains the customer. We call these 36 messages
\emph{brand-dependent}. Together they make up 81 rows, and most of them describe a switch between two
providers. Appendix~\ref{app:examples} shows further examples.

Table~\ref{tab:data} summarises the data. The released file has 317 rows and 231 distinct messages,
and 125 rows (39.4\%) are churn. \textcite{abbet-2018-churn} report 307 English messages; we use the
file as it was released. We strip surrounding white space from the messages and merge empty brand
cells with the tag \texttt{no\_brand}. No row and no label is changed.

\begin{table}[!ht]
\centering
\footnotesize
\begin{threeparttable}
\caption{Rows and churn rate of the English chatbot data}
\label{tab:data}
\begin{tabular}{lrrr}
\toprule
 & Rows & Churn & Share churn \\
\midrule
All rows & 317 & 125 & 39.4\% \\
\addlinespace
\multicolumn{4}{l}{\textit{By type of message}} \\
\quad Brand-dependent (36 messages) & 81 & 39 & 48.1\% \\
\quad Other (195 messages) & 236 & 86 & 36.4\% \\
\addlinespace
\multicolumn{4}{l}{\textit{By brand}} \\
\quad Swisscom & 116 & 57 & 49.1\% \\
\quad Salt & 46 & 12 & 26.1\% \\
\quad Vodafone & 13 & 6 & 46.2\% \\
\quad Orange, Sunrise & 16 & 7 & 43.8\% \\
\quad O2, UPC, Verizon, SFR & 10 & 2 & 20.0\% \\
\quad No brand & 116 & 41 & 35.3\% \\
\bottomrule
\end{tabular}
\tabnote{The 317 rows contain 231 distinct messages with 8.4 words on average (maximum 27).}
\end{threeparttable}
\end{table}

\subsection{Task Definition}

All systems receive the same task definition (Appendix~\ref{app:prompts}). It names the brand and
states the labelling rule explicitly. A message is churn for a brand if the customer intends to
leave, cancel or switch away from it, or is seriously considering doing so. Switching to the brand
is not churn. We state the rule because unstated labelling rules are the most common reason for
errors of TDMs \parencite{suppa-2026-do}. Each system returns a probability $p_i$ that row $i$ is
churn, and the predicted label is churn if $p_i > 0.5$.

\subsection{Systems}

Table~\ref{tab:systems} lists the 25 systems, and Appendix~\ref{app:models} gives their identifiers
and prices.

\begin{table}[!ht]
\footnotesize
\caption{Evaluated systems}
\label{tab:systems}
\begin{minipage}[t]{0.53\textwidth}
\textit{Typed decision models}\\[2pt]
\begin{tabular}[t]{@{}lll@{}}
\toprule
Model & Vendor & Weights \\
\midrule
Jev 1.14 & TypeSafe & API \\
Microsoft-Decision-1.1 & Microsoft & API \\

GPT-6 Luna Decisions & OpenAI & API \\
Decider V1.3 & Perplexity & API \\
d1.2-ultra & Liquid AI & API \\
Drex Max & Nace.AI & API \\
Solar Decide & Upstage & API \\
Classify-1 & ScaleDown & API \\
Reflex & Morph & API \\
HeyGen Decision 1 & HeyGen & API \\
Clef & Cloudflare & open (Qwen3.8-27B) \\
Clef Flash & Cloudflare & open (Qwen3.5-9B) \\
Mammoth-1 & Baseten & open (Qwen3.5-27B) \\
Elephant-1 & Baseten & open (Qwen3.5-9B) \\
Span-02 & Respan & open (Gemma 4 26B-A4B) \\
Kev-4B & Jared Palmer & open (Qwen3.5-4B) \\
\bottomrule
\end{tabular}
\end{minipage}\hfill
\begin{minipage}[t]{0.43\textwidth}
\textit{Prompted LLMs}\\[2pt]
\begin{tabular}[t]{@{}lll@{}}
\toprule
Model & Vendor & Weights \\
\midrule
Claude Sonnet 5.5 & Anthropic & API \\
Claude Haiku 5.5 & Anthropic & API \\
GPT-6.2 Sol & OpenAI & API \\
GPT-6 Luna & OpenAI & API \\
Qwen3.8-27B & Qwen & open \\
Qwen3.5-27B & Qwen & open \\
Qwen3.5-9B & Qwen & open \\
\bottomrule
\end{tabular}

\vspace{10pt}
\textit{Trained classifiers}\\[2pt]
\begin{tabular}[t]{@{}ll@{}}
\toprule
Model & Features \\
\midrule
TF-IDF + LR & words and word pairs \\
Embeddings + LR & sentence embeddings \\
\bottomrule
\end{tabular}
\end{minipage}
\tabnote{API: weights not published, model available only through the vendor's interface. Open:
weights published, base model in parentheses. LR: logistic regression. All typed decision models and LLMs are called through OpenRouter; the
trained classifiers run locally. Gemini Decision 1 meets the selection rule but rejected all requests
(Section~3.3).}
\end{table}

\textbf{Typed decision models.} We include the flagship TDM of each vendor that is available on
OpenRouter's Decisions API as of mid-November 2026. The flagship is the vendor's largest or most
capable general-purpose model for text, in its latest version \parencite{openrouter-2026-models}. We
exclude older versions (for example Jev 1.13 and Decider V1.2), aliases that point to the latest
version, and smaller or faster variants (Decider Flash, d1.2, Solar Decide Flash, HeyGen Decision
Turbo, Joey-1 and Clef Omni). We also exclude one model that is tuned for finance (Drex Max Finance),
one that searches the web before it answers (Parallel Decisions Fast), two models for behaviour
scoring and for extraction (Span-01 and Clairic), and one experimental model (Tev1 4B Experimental).
Gemini Decision 1 from Google also meets the rule. OpenRouter rejected every request to it,
however, including a simple test request unrelated to churn, with the message that the request
violated the provider's terms of service. We could therefore not evaluate it, which leaves 16 TDMs.
We then add Baseten Elephant-1 and Cloudflare Clef Flash, two smaller models that are fine-tuned
from Qwen3.5-9B. With them, every open base model in our set of LLMs has a decision-tuned
counterpart. Each row is sent as one request. Its state is the same text that the LLMs receive as the user
message, the brand and the message, and a single yes/no question carries the task definition. We use the probability of ``yes'' as $p_i$. We
do not use the confidence field of the response. It measures how concentrated the probabilities are,
which is different from how likely the answer is to be correct \parencite{typesafe-2026-confidence}.

\textbf{Prompted LLMs.} We include a large and a small closed model from Anthropic and from OpenAI,
which are the models that a company would most likely use today. GPT-6 Luna is also offered as a
TDM, so the same model can be compared in both forms. We add three open-weight models that a company
can host itself. Qwen3.8-27B is the base model of Clef, Qwen3.5-27B of Mammoth-1, and Qwen3.5-9B of
Elephant-1 and Clef Flash. Each row is one chat request with the task definition as the system
prompt. The model answers in JSON with the probability $p_i$ that the message is churn for the given
brand, a number between 0 and 1. We request this format with a fixed JSON schema (structured
outputs), which all seven models support. The closed models do not accept a temperature setting; for
the Qwen models, we set the temperature to 0. The output limit is 2,000 tokens.

We run every LLM at its lowest reasoning setting. Reasoning can be switched off for Claude Haiku 5.5,
GPT-6 Luna and the three Qwen models. Claude Sonnet 5.5 and GPT-6.2 Sol always reason, so we use
their lowest effort (``low''). There are three reasons
for this choice. First, recognising churn in a short message is the kind of fast, routine judgement
for which TDMs are built, so the fair comparison is with an LLM that also answers directly. Second, a
chatbot has to answer while the customer is waiting, and reasoning models spend many tokens even on
easy inputs \parencite{chen-2025-do}, which raises both latency and cost. Third, earlier comparisons
of TDMs with fast closed LLMs made the same choice \parencite{chen-2026-system1arena}. We record the
reasoning tokens that each model reports, so that any remaining reasoning is visible in the results.

\textbf{Trained classifiers.} We train two classifiers on the labelled rows. The first uses TF-IDF features of single words and word pairs, with
sublinear term frequencies and the brand added as an extra token \parencite{salton-1988-term}. The second uses 384-dimensional sentence embeddings of the brand and
the message, computed with the all-MiniLM-L6-v2 encoder \parencite{reimers-2019-sentencebert,wang-2020-minilm}.
Both use logistic regression with the default settings of scikit-learn, that is, L2 regularisation
with $C = 1$ and equal weights for the two classes \parencite{pedregosa-2011-scikitlearn}. We keep the
weights equal because reweighting the classes would shift the probabilities away from the observed
churn rate. We evaluate them with five-fold cross-validation grouped by
message, so that all brand variants of a message fall into the same fold. Every row thus receives a
prediction from a model that has not seen the message. The hyperparameters are fixed in advance.

\subsection{Evaluation Measures}

\textbf{Discrimination.} We report precision, recall and F1 for the class churn at the threshold
$p_i > 0.5$, and the area under the ROC curve (AUROC), which does not depend on a threshold.

\textbf{Calibration.} For each row, the confidence of the decision is $\hat{c}_i = \max(p_i, 1 - p_i)$.
We sort the rows into $B = 10$ bins of equal width on $[0.5, 1]$ and compute the expected
calibration error \parencite{naeini-2015-obtaining,guo-2017-calibration},
\begin{equation}
  \mathrm{ECE} = \sum_{b=1}^{B} \frac{n_b}{N} \left| \mathrm{acc}(b) - \mathrm{conf}(b) \right|,
  \label{eq:ece}
\end{equation}
where $n_b$ is the number of rows in bin $b$, $N = 317$, and $\mathrm{acc}(b)$ and
$\mathrm{conf}(b)$ are the accuracy and the mean confidence in that bin. We also report the Brier
score, $\frac{1}{N} \sum_i (p_i - y_i)^2$ with $y_i \in \{0, 1\}$ the label
\parencite{brier-1950-verification}, and show reliability diagrams with the same bins.

\textbf{Automation.} A company that acts automatically only on confident decisions passes the
remaining messages to a person. We sort the rows by $\hat{c}_i$ and compute the accuracy among the
most confident share of rows, for shares from 10\% to 100\%. From this curve we read the largest
share of messages that can be decided automatically while the accuracy among them stays at or above
95\%. These values are descriptive. On new messages, a threshold that is chosen on the same data
usually reaches a lower accuracy than these values suggest \parencite{brom-2026-confidence}.

\textbf{Latency and cost.} Latency is the wall-clock time of each successful request, measured at
the client; we report its median and 95th percentile. For the trained classifiers, it is the time
to score one row on a CPU. Cost is the price that OpenRouter bills for each request, reported per
1,000 decisions. The trained classifiers have no cost per decision.

\textbf{Intervals.} Because a message occurs with several brands, rows are not independent. We
compute 95\% intervals with a cluster bootstrap that resamples messages, with 2,000 resamples
\parencite{field-2007-bootstrapping}.

\subsection{Statistical Analysis}

The mixed-effects models use the correctness of each decision as the outcome. Let $y_{si} = 1$ if
system $s$ decides row $i$ correctly, let $d_i = 1$ if row $i$ belongs to a brand-dependent
message, and let $m(i)$ be the message of row $i$. The first model compares the three families of
systems, $f(s) \in \{\text{TDM}, \text{LLM}, \text{trained}\}$:
\begin{equation}
  \operatorname{logit} P(y_{si} = 1) = \beta_{f(s)} + \gamma_{f(s)} d_i + u_{m(i)} + v_s + w_s d_i,
  \label{eq:glmm1}
\end{equation}
with a random intercept $u_{m(i)} \sim N(0, \sigma_u^2)$ for each message, and a random intercept
$v_s$ and random slope $w_s$ for each system, with a joint normal distribution. The coefficients
$\gamma_{f(s)}$ measure how much the accuracy of each family changes on brand-dependent messages.
With random effects for the systems, the models of a family are treated as a sample from that
family, as recommended for designs in which items and conditions are crossed
\parencite{baayen-2008-mixed,barr-2013-random}. The second model replaces the families with the
individual systems:
\begin{equation}
  \operatorname{logit} P(y_{si} = 1) = \beta_s + \gamma_s d_i + u_{m(i)}.
  \label{eq:glmm2}
\end{equation}
From the second model, we test planned contrasts. The first group compares each open TDM with the
LLM from which it was fine-tuned: Clef with Qwen3.8-27B, Mammoth-1 with Qwen3.5-27B, and Elephant-1
and Clef Flash with Qwen3.5-9B. Further contrasts compare GPT-6 Luna Decisions with GPT-6 Luna,
Elephant-1 with Mammoth-1, and Jev with each closed LLM. We fit both models with \texttt{glmer} from the R package lme4
\parencite{bates-2015-fitting} and convert the results into probabilities of a correct decision with
emmeans \parencite{lenth-2026-emmeans}. The $p$-values of the contrasts are adjusted for multiple
comparisons within each group of contrasts. If a model does not converge, for example because a
system makes almost no errors, we fit the same model in brms with weakly informative priors
\parencite{burkner-2017-brms}.

\subsection{Implementation}

All TDMs and LLMs are called through OpenRouter: the LLMs through its chat interface and the TDMs
through its Decisions API, which is in an alpha stage \parencite{openrouter-2026-decisions}. Models are
fixed by their version identifier, and we record the model and provider that answered each request.
Up to eight requests run in parallel, and requests that fail because of rate limits or server errors
are repeated. An answer is invalid if it cannot be parsed, if its probability lies outside $[0, 1]$,
if the model refuses to answer, or if the answer is missing. Invalid answers receive $p_i = 0.5$,
which counts as the prediction no churn, so that every system is evaluated on all 317 rows.
All requests were sent on 19 November 2026. The trained classifiers ran on the CPU of a laptop
(Intel Core i7-1165G7).

%% file: sections/04_results.tex
\section{Results}
\label{sec:results}

\subsection{Accuracy and Calibration}

Table~\ref{tab:results} reports the main measures for all 25 systems. Six TDMs form a leading group:
Decider V1.3, d1.2-ultra, Drex Max, Microsoft-Decision-1.1, Jev 1.14 and HeyGen Decision 1. They
reach F1 scores between 0.93 and 0.95 and an AUROC of 0.99, and their expected calibration error
lies between 0.02 and 0.07. The best LLM, GPT-6.2 Sol, is in the same range (F1 0.94, AUROC 0.99,
ECE 0.03). The other closed LLMs reach F1 scores of about 0.90. The TDMs differ much more among
themselves than the LLMs do. The three smallest open TDMs, Kev-4B, Clef Flash and Elephant-1, reach
F1 scores of only 0.73 to 0.81, and Solar Decide reaches 0.83. Across their members, the two families end up close. The median F1 is 0.89 for the TDMs and 0.90
for the LLMs, and the median ECE is 0.05 and 0.06. As a class, the TDMs are therefore not better calibrated
than the LLMs, although their best members are among the best calibrated systems. The two trained
classifiers are clearly weaker, with F1 scores of 0.62 and 0.65 and ECE values of 0.11 and 0.14.

\begingroup
\footnotesize
\setlength{\tabcolsep}{4pt}
\setlength{\LTleft}{0pt}
\setlength{\LTright}{0pt}
\setlength{\LTcapwidth}{\textwidth}
\setlength{\LTpre}{8pt}
\setlength{\LTpost}{0pt}
\makeatletter
\begin{longtable}{@{\extracolsep{\fill}}lccccccr@{}}
\caption{Results of all systems on the 317 rows}
\label{tab:results} \\
\toprule
System & F1 & AUROC & ECE & Automated & Brand-dep. & Latency & US\$/1,000 \\
\midrule
\endfirsthead
\multicolumn{8}{@{}l}{\textit{Table~\ref{tab:results}, continued}} \\
\toprule
System & F1 & AUROC & ECE & Automated & Brand-dep. & Latency & US\$/1,000 \\
\midrule
\endhead
\bottomrule
\endlastfoot
\input{tables/results_rows.tex}
\end{longtable}
\makeatother
\vspace{-6pt}
\tabnote{Values in brackets are 95\% intervals from a cluster bootstrap over messages. Automated: largest
share of rows that can be decided automatically with an accuracy of at least 95\%, when the most
confident rows are decided first. Brand-dep.: accuracy on the 81 rows of brand-dependent messages.
Latency: median time per decision. Within each family, systems are sorted by F1.}
\endgroup
\vspace{6pt}

Nine of the 7,925 decisions were invalid. Qwen3.5-27B returned malformed numbers eight times, for
example a long run of digits instead of a probability, and GPT-6.2 Sol refused to answer the message
``I hate swisscom, I quit'' once. All other systems answered every row. Neither of the two LLMs that
cannot switch off reasoning used reasoning tokens at the lowest effort; all LLMs answered with 11 to
18 output tokens on average.

\subsection{Automation, Speed and Cost}

\begin{figure}[H]
\caption{Accuracy among the messages that are decided automatically}
\label{fig:coverage}
\centering
\includegraphics[width=\textwidth]{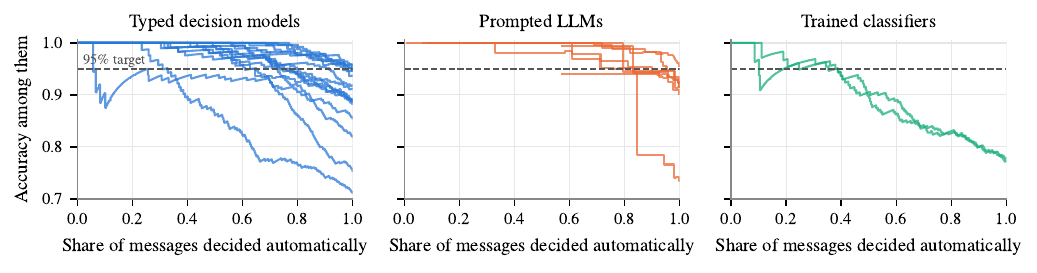}
\tabnote{Each line is one system. Messages are decided automatically in order of decreasing
confidence; the dashed line marks an accuracy of 95\%.}
\end{figure}

Figure~\ref{fig:coverage} shows how accuracy changes when a company decides only the most confident
messages automatically and passes the rest to staff. For the six leading TDMs and for GPT-6.2 Sol,
accuracy stays at or above 95\% almost to the end: they can decide 96 to 100\% of the rows
automatically at this level. Claude Sonnet 5.5 reaches 94\%. Several systems with a good F1 score
automate much less, because a few confident errors already push their accuracy below 95\%; Qwen3.8-27B,
for example, reaches 59\%. The intervals of this measure are wide (Table~\ref{tab:results}), so small
differences between systems should not be read as a ranking. The trained classifiers reach 95\%
accuracy only on the most confident 38\% of the rows.

\begin{figure}[!ht]
\caption{Quality against cost per 1,000 decisions}
\label{fig:cost}
\centering
\includegraphics[width=\textwidth]{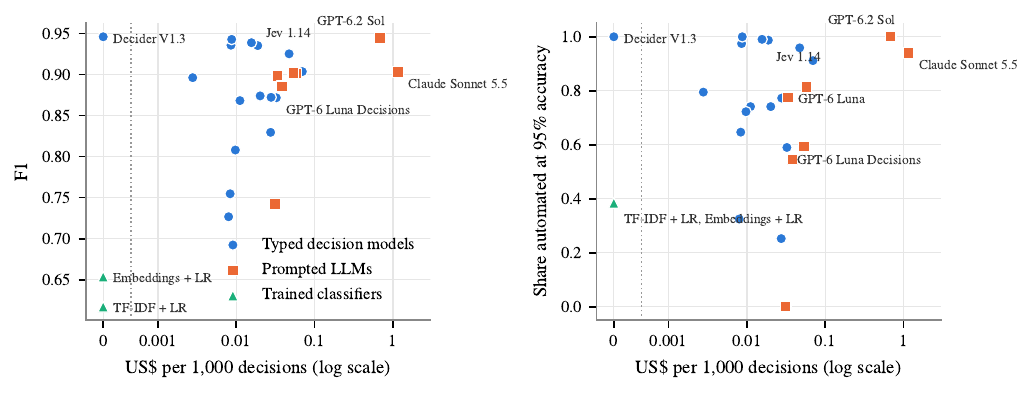}
\tabnote{Left: F1. Right: share of rows that can be decided automatically at 95\% accuracy. Systems
without cost per decision (the trained classifiers and the free Decider V1.3) are drawn at zero,
left of the dotted line.}
\end{figure}

The TDMs are faster and cheaper (Figure~\ref{fig:cost}). The leading TDMs answer in a median of 0.26
to 0.60 seconds, measured at the client and including the gateway. The closed LLMs need 0.95 to 1.51
seconds, and Solar Decide is an exception among the TDMs, with a 95th percentile of 7.5 seconds. The
TDMs cost between 0 and 0.07 US dollars per 1,000 decisions, against 0.03 to 1.16 dollars for the
LLMs. d1.2-ultra matches GPT-6.2 Sol in F1 and in the share of automated rows at about one
seventy-eighth of its cost (0.009 against 0.68 dollars) and one fifth of its latency. Jev costs 0.019
dollars per 1,000 decisions, about one sixtieth of Claude Sonnet 5.5. The trained classifiers have
no cost per decision and answer in milliseconds on a laptop CPU.

\subsection{Brand-Dependent Messages and Base Models}

Table~\ref{tab:glmm} summarises the mixed-effects models. The probabilities refer to a typical
message, that is, a message with an average random effect, and are therefore higher than the raw
accuracies in Table~\ref{tab:results}. The random intercept of the messages has a variance of 2.38
on the logit scale, which shows that some messages are much harder than others for all systems.

On other messages, the three families decide a typical message correctly with a probability of 0.97
(TDMs), 0.96 (LLMs) and 0.94 (trained classifiers). On brand-dependent messages, the probability
changes by $-0.03$ for the TDMs (95\% interval $-0.07$ to $0.02$) and by $0.00$ for the LLMs ($-0.04$ to $0.03$). For the trained classifiers, it changes by $-0.37$
($-0.76$ to $0.03$). The difference
in this change between TDMs and LLMs is small and not significant ($-0.02$, adjusted $p = 0.25$).
The interval for the trained classifiers is wide because the family has only two members, but their
raw accuracy on brand-dependent rows (51 and 60\%) is close to guessing.

The comparison of open TDMs with the LLMs from which they were fine-tuned depends on the type of
message (Table~\ref{tab:glmm}, lower part). On other messages, decision tuning helps the two small
models: Elephant-1 is more likely to be correct than Qwen3.5-9B by 0.13 (0.04 to 0.21) and Clef Flash
by 0.10 (0.01 to 0.18). The larger TDMs and GPT-6 Luna Decisions do not differ from their base
models. On brand-dependent messages, all five TDMs are less likely to be correct than their base
models. The difference is largest for Clef Flash ($-0.25$, $-0.44$ to $-0.05$) and Mammoth-1
($-0.17$, $-0.32$ to $-0.02$); for the other three, the intervals include zero. In raw accuracy on
brand-dependent rows, Mammoth-1 reaches 77\% and its base model Qwen3.5-27B 96\%. The remaining
planned contrasts show no difference between Jev and any closed LLM (all adjusted $p \geq 0.21$),
and a lower probability for Elephant-1 than for the larger Mammoth-1 of the same vendor ($-0.05$,
$-0.08$ to $-0.02$).

\begin{table}[!ht]
\centering
\footnotesize
\begin{threeparttable}
\caption{Mixed-effects models: probability of a correct decision on a typical message}
\label{tab:glmm}
\begin{tabular}{@{}lccc@{}}
\toprule
 & Other messages & Brand-dependent messages & Change \\
\midrule
\multicolumn{4}{@{}l}{\textit{Model 1: families}} \\
Typed decision models & 0.97 [0.95, 0.98] & 0.94 [0.88, 0.97] & $-0.03$ [$-0.07$, 0.02] \\
Prompted LLMs & 0.96 [0.94, 0.98] & 0.96 [0.89, 0.99] & 0.00 [$-0.04$, 0.03] \\
Trained classifiers & 0.94 [0.85, 0.98] & 0.57 [0.19, 0.88] & $-0.37$ [$-0.76$, 0.03] \\
\addlinespace
\multicolumn{4}{@{}l}{\textit{Model 2: open TDM minus its base LLM}} \\
Clef $-$ Qwen3.8-27B & 0.01 [$-0.02$, 0.04] & $-0.03$ [$-0.09$, 0.03] & \\
Mammoth-1 $-$ Qwen3.5-27B & 0.02 [$-0.01$, 0.05] & $-0.17$ [$-0.32$, $-0.02$] & \\
Elephant-1 $-$ Qwen3.5-9B & 0.13 [0.04, 0.21] & $-0.08$ [$-0.24$, 0.07] & \\
Clef Flash $-$ Qwen3.5-9B & 0.10 [0.01, 0.18] & $-0.25$ [$-0.44$, $-0.05$] & \\
Luna Decisions $-$ GPT-6 Luna & 0.00 [$-0.03$, 0.02] & $-0.08$ [$-0.20$, 0.03] & \\
\bottomrule
\end{tabular}
\tabnote{Probabilities and differences in probability with 95\% intervals, from Eq.~(\ref{eq:glmm1})
(Model 1) and Eq.~(\ref{eq:glmm2}) (Model 2). A typical message has a random effect of zero. The
intervals of Model 2 are adjusted for the five comparisons within each type of message.}
\end{threeparttable}
\end{table}

%% file: tables/results_rows.tex
\multicolumn{8}{@{}l}{\textit{Typed decision models}} \\
Decider V1.3 & 0.95 [0.91, 0.97] & 0.99 [0.98, 1.00] & 0.04 [0.02, 0.06] & 100\% & 98\% & 0.41 s & 0.000 \\
d1.2-ultra & 0.94 [0.91, 0.97] & 0.99 [0.99, 1.00] & 0.03 [0.02, 0.06] & 100\% & 99\% & 0.30 s & 0.009 \\
Drex Max & 0.94 [0.90, 0.97] & 0.99 [0.98, 1.00] & 0.02 [0.02, 0.06] & 99\% & 99\% & 0.60 s & 0.016 \\
Microsoft-Decision-1.1 & 0.94 [0.90, 0.96] & 0.99 [0.99, 1.00] & 0.04 [0.04, 0.07] & 97\% & 99\% & 0.38 s & 0.009 \\
Jev 1.14 & 0.94 [0.90, 0.97] & 0.99 [0.98, 1.00] & 0.07 [0.06, 0.10] & 99\% & 99\% & 0.26 s & 0.019 \\
HeyGen Decision 1 & 0.93 [0.89, 0.96] & 0.99 [0.98, 1.00] & 0.02 [0.02, 0.06] & 96\% & 99\% & 0.35 s & 0.047 \\
Clef & 0.90 [0.86, 0.94] & 0.98 [0.96, 0.99] & 0.04 [0.03, 0.08] & 91\% & 94\% & 0.39 s & 0.070 \\
Span-02 & 0.90 [0.85, 0.93] & 0.98 [0.97, 0.99] & 0.05 [0.04, 0.09] & 79\% & 95\% & 0.50 s & 0.003 \\
Classify-1 & 0.87 [0.83, 0.91] & 0.97 [0.96, 0.99] & 0.10 [0.08, 0.14] & 74\% & 83\% & 0.43 s & 0.020 \\
Mammoth-1 & 0.87 [0.83, 0.91] & 0.98 [0.97, 0.99] & 0.05 [0.04, 0.09] & 77\% & 77\% & 0.28 s & 0.028 \\
GPT-6 Luna Decisions & 0.87 [0.82, 0.91] & 0.95 [0.93, 0.97] & 0.06 [0.05, 0.11] & 59\% & 83\% & 0.39 s & 0.032 \\
Reflex & 0.87 [0.82, 0.91] & 0.99 [0.98, 0.99] & 0.04 [0.03, 0.09] & 74\% & 95\% & 0.30 s & 0.011 \\
Solar Decide & 0.83 [0.78, 0.88] & 0.93 [0.90, 0.95] & 0.08 [0.05, 0.12] & 25\% & 75\% & 1.30 s & 0.027 \\
Elephant-1 & 0.81 [0.75, 0.86] & 0.97 [0.95, 0.99] & 0.06 [0.04, 0.11] & 72\% & 75\% & 0.26 s & 0.010 \\
Clef Flash & 0.75 [0.69, 0.81] & 0.96 [0.94, 0.98] & 0.12 [0.09, 0.17] & 65\% & 60\% & 0.36 s & 0.008 \\
Kev-4B & 0.73 [0.67, 0.78] & 0.92 [0.89, 0.95] & 0.10 [0.08, 0.16] & 32\% & 62\% & 0.66 s & 0.008 \\
\addlinespace
\multicolumn{8}{@{}l}{\textit{Prompted LLMs}} \\
GPT-6.2 Sol & 0.94 [0.91, 0.97] & 0.99 [0.98, 1.00] & 0.03 [0.02, 0.05] & 100\% & 94\% & 1.50 s & 0.684 \\
Claude Sonnet 5.5 & 0.90 [0.85, 0.95] & 0.99 [0.97, 0.99] & 0.07 [0.05, 0.09] & 94\% & 94\% & 1.09 s & 1.164 \\
Qwen3.8-27B & 0.90 [0.86, 0.94] & 0.98 [0.96, 0.99] & 0.07 [0.04, 0.11] & 59\% & 98\% & 0.56 s & 0.053 \\
Claude Haiku 5.5 & 0.90 [0.86, 0.94] & 0.97 [0.95, 0.99] & 0.03 [0.03, 0.07] & 81\% & 94\% & 0.95 s & 0.058 \\
GPT-6 Luna & 0.90 [0.85, 0.94] & 0.96 [0.94, 0.98] & 0.06 [0.03, 0.09] & 78\% & 93\% & 1.51 s & 0.033 \\
Qwen3.5-27B & 0.89 [0.84, 0.93] & 0.98 [0.96, 0.99] & 0.05 [0.03, 0.09] & 55\% & 96\% & 2.54 s & 0.038 \\
Qwen3.5-9B & 0.74 [0.68, 0.80] & 0.91 [0.87, 0.94] & 0.24 [0.18, 0.29] & 0\% & 84\% & 0.57 s & 0.031 \\
\addlinespace
\multicolumn{8}{@{}l}{\textit{Trained classifiers}} \\
Embeddings + LR & 0.65 [0.56, 0.73] & 0.86 [0.82, 0.90] & 0.11 [0.09, 0.16] & 38\% & 51\% & 9.0 ms & -- \\
TF-IDF + LR & 0.62 [0.52, 0.70] & 0.88 [0.83, 0.92] & 0.14 [0.11, 0.19] & 38\% & 60\% & 0.8 ms & -- \\

%% file: sections/05_discussion.tex
\section{Discussion and Conclusion}
\label{sec:discussion}

The results answer the research question with a qualified yes. The best TDMs detect churn intent as well as the best LLM in our set, and their probabilities are
well calibrated. They allow almost all messages of this corpus to be decided automatically at 95\%
accuracy. They do this in a few hundred
milliseconds and for a few cents per 1,000 decisions. For a company, this changes the economics of the task. At these prices, it costs almost nothing to
check every chatbot message for churn intent, and the latency is low enough to answer while the
customer is still in the conversation.

The qualification is that TDMs differ widely. The six leading models come from six different vendors, while the smallest open models fall far
behind. A company should therefore not
choose ``a TDM'' but a particular one, and test it on its own messages first. The vendors' claims hold in part. Jev answers in a median of 0.26 seconds, within the 70 to 500
milliseconds that TypeSafe advertises. The leading paid TDMs are 14 to 78 times cheaper than GPT-6.2
Sol, the most accurate LLM, which is far below the factor of 444 on TypeSafe's home page. Their calibration is good, but as a class it is not better than that of the LLMs. On this task, a probability stated by a modern LLM is about as reliable as the probability of a
TDM.

The messages whose label depends on the brand separate the trained classifiers from the other two
families. A classifier that only counts words cannot tell who gains and who loses the customer, and
it decides these messages close to chance. TDMs and LLMs handle them about as well as other
messages. The comparison of open TDMs with their base models adds a caveat. Decision tuning helped the small
open models on ordinary messages, but every open TDM was less accurate than its own base model on
brand-dependent messages, clearly so for Mammoth-1 and Clef Flash. Tuning a model for fast typed
decisions seems to cost some of its ability to work out whose customer is leaving. This agrees with two earlier observations: TDMs follow the wording of a question literally, and
they fail most often on labelling rules that the label itself does not state \parencite{suppa-2026-do,typesafe-2026-model}.

The open models matter for companies that cannot send customer messages to an external service.
Open TDMs and open LLMs such as Qwen3.8-27B can be hosted on a company's own hardware. In our
results, the open LLM of 27 billion parameters reached an F1 score of 0.90, close to the closed
models, whereas the open TDMs were mostly weaker than the leading closed TDMs. Whether self-hosting
is cheaper than an API depends on the volume of messages and on how well the hardware is used
\parencite{irugalbandara-2024-scaling}.

\textbf{Limitations.} The messages were written by crowd workers in a game-like setting and are not
real customer traffic, and the corpus is small: 317 rows from 231 messages. The corpus has been
public since 2018, so some models may have seen it during training, and the very high accuracy of
the best systems may partly reflect this. Each system was run once, through one gateway, on one day
in November 2026, so latency includes the network and the provider's load, and prices and model
versions may change. The share of messages that can be automated was computed on the same data that
define it; a threshold chosen on these rows would reach a lower accuracy on new messages
\parencite{brom-2026-confidence}. Finally, we studied English messages only.

\textbf{Conclusion.} Typed decision models are a practical option for detecting churn intent in
customer chatbots. The best of them are as accurate and as well calibrated as the best prompted LLM, at a fraction of
its cost and latency. They also handle messages whose meaning depends on the brand far better than
classifiers trained on a few hundred examples. Their quality, however, depends
strongly on the vendor and the size of the model. Companies should test a candidate on their own
messages, including messages that name competitors, before they let it decide automatically.

%% file: sections/appendix.tex
\section{Examples from the Data}
\label{app:examples}

Table~\ref{tab:examples} shows rows from the English chatbot data of \textcite{abbet-2018-churn},
written as they appear in the file. The first block shows brand-dependent messages, whose label
changes with the brand. The second block shows messages with a single brand or without one,
including complaints without an intention to leave and an example of the noise that the game-like
collection produced.

\begin{table}[H]
\centering
\footnotesize
\begin{threeparttable}
\caption{Examples of rows from the data}
\label{tab:examples}
\begin{tabular}{@{}p{9.2cm}ll@{}}
\toprule
Message & Brand & Label \\
\midrule
\multicolumn{3}{@{}l}{\textit{Brand-dependent messages}} \\
I want to switch from Swisscom to salt & Swisscom & churn \\
 & Salt & no churn \\
I switch from Vodafone to Swisscom & Vodafone & churn \\
 & Swisscom & no churn \\
I am leaving salt.... Is swisscom or sunrise a good alternative ? & Salt & churn \\
 & Swisscom, Sunrise & no churn \\
I'm going to swisscom & no brand & churn \\
 & Swisscom & no churn \\
\addlinespace
\multicolumn{3}{@{}l}{\textit{Other messages}} \\
Swisscom service is the worst, I'm leaving next month & Swisscom & churn \\
I want to terminate my contract to salt & Salt & churn \\
I quit & no brand & churn \\
I want to know more infos about Swisscom & no brand & churn \\
I'm not a fan of swisscom & Swisscom, no brand & no churn \\
I am fed up with you & no brand & no churn \\
I should change contract option. What are the offers ? & no brand & no churn \\
CRAAAAAAAAB people CRAAAAAAAAA people & no brand & no churn \\
\bottomrule
\end{tabular}
\tabnote{Rows without a brand are judged from the perspective of the customer's current provider,
which the message does not name. Interest in another operator therefore counts as churn.}
\end{threeparttable}
\end{table}

\section{Task Definition, Prompts and Decision Request}
\label{app:prompts}

All systems receive the same task definition. The prompted LLMs receive it as the system prompt,
followed by the answer format, and the brand and the message as the user message. The TDMs receive the same user message as their state and the task definition as the instructions
of a single yes/no question.

\begin{sysprompt}[Task definition (all systems)]
You classify short customer messages for churn intent with respect to a given brand.

Imagine that the given brand receives the message. The message is churn if it shows that the customer intends to leave, cancel, or switch away from that brand, or is seriously considering doing so. Otherwise it is not churn. This includes switching or moving to the brand, complaints that do not express an intention to leave, questions, and unrelated content. The same wording can be churn for one brand and not for another.

If the brand is "no_brand", no brand is given. Judge the message from the perspective of the customer's current provider: the message is churn if the customer intends to leave or switch providers, including switching to or seriously considering a named competitor.
\end{sysprompt}

\begin{sysprompt}[Answer format (prompted LLMs, appended to the task definition)]
Respond with only a JSON object of the form {"churn_probability": p}, where p is a number between 0 and 1 that gives the probability that the message is churn for the given brand. No other text.
\end{sysprompt}

\begin{userprompt}[User message (prompted LLMs)]
brand: <brand>
message: "<message>"
\end{userprompt}

\begin{userprompt}[Request to the Decisions API (typed decision models)]
{
  "model": "<model identifier>",
  "state": "brand: <brand>\nmessage: \"<message>\"",
  "questions": {
    "churn": {
      "type": "noul",
      "instructions": "<task definition> Is the message churn for the given brand?"
    }
  }
}
\end{userprompt}

The response of a TDM contains a probability for ``yes'' and ``no'' for the question \texttt{churn}.
We use the probability of ``yes'' as the probability of churn. Appendix~\ref{app:io} shows complete
requests and responses.

\section{Example Requests and Responses}
\label{app:io}

Table~\ref{tab:io} shows the raw outputs of one system of each kind for a brand-dependent message,
paired with both of its brands, and for one other message. The TDMs return the probability of
``yes'' in the field \texttt{noul}; the LLMs return the JSON object that the answer format asks for.
The trained classifier has no text output, only a probability.

\begin{table}[H]
\centering
\footnotesize
\begin{threeparttable}
\caption{Raw outputs for three rows}
\label{tab:io}
\begin{tabular}{@{}ll@{}}
\toprule
System & Raw output \\
\midrule
\multicolumn{2}{@{}l}{\textit{``I want to switch from Swisscom to salt'', brand Swisscom (label: churn)}} \\
Jev 1.14 & \texttt{\{"churn": \{"type": "noul", "noul": 0.99\}\}} \\
Mammoth-1 & \texttt{\{"churn": \{"type": "noul", "noul": 0.9446920884142371\}\}} \\
Claude Sonnet 5.5 & \texttt{\{"churn\_probability": 0.98\}} \\
Qwen3.5-9B & \texttt{\{"churn\_probability": 1.0\}} \\
TF-IDF + LR & probability 0.64 \\
\addlinespace
\multicolumn{2}{@{}l}{\textit{``I want to switch from Swisscom to salt'', brand Salt (label: no churn)}} \\
Jev 1.14 & \texttt{\{"churn": \{"type": "noul", "noul": 0.07\}\}} \\
Mammoth-1 & \texttt{\{"churn": \{"type": "noul", "noul": 0.13144888273788094\}\}} \\
Claude Sonnet 5.5 & \texttt{\{"churn\_probability": 0.02\}} \\
Qwen3.5-9B & \texttt{\{"churn\_probability": 0.0\}} \\
TF-IDF + LR & probability 0.50 \\
\addlinespace
\multicolumn{2}{@{}l}{\textit{``I am leaving salt'', brand Salt (label: churn)}} \\
Jev 1.14 & \texttt{\{"churn": \{"type": "noul", "noul": 0.97\}\}} \\
Mammoth-1 & \texttt{\{"churn": \{"type": "noul", "noul": 0.9722769855937997\}\}} \\
Claude Sonnet 5.5 & \texttt{\{"churn\_probability": 0.97\}} \\
Qwen3.5-9B & \texttt{\{"churn\_probability": 1.0\}} \\
TF-IDF + LR & probability 0.57 \\
\bottomrule
\end{tabular}
\tabnote{All outputs of all systems are in the released prediction files.}
\end{threeparttable}
\end{table}

\clearpage
\section{Models, Identifiers and Prices}
\label{app:models}

Table~\ref{tab:models} lists the OpenRouter identifiers of all models with their list prices on
19 November 2026. The listing date of a TDM is the date on which OpenRouter first listed it.
OpenRouter reported a dated snapshot for every TDM except Clef and Clef Flash, for example
\texttt{typesafe/jev-1.14-20261014}; the version that answered is stored with every prediction.
Requests to the open models were served by several providers, which OpenRouter chose per request.

\begin{table}[H]
\centering
\footnotesize
\begin{threeparttable}
\caption{OpenRouter identifiers and list prices}
\label{tab:models}
\begin{tabular}{@{}llrr@{}}
\toprule
Identifier & Listed & \multicolumn{2}{c@{}}{US\$ per million tokens} \\
\cmidrule(l){3-4}
 & & Input & Output \\
\midrule
\multicolumn{4}{@{}l}{\textit{Typed decision models}} \\
\texttt{typesafe/jev-1.14} & 2026-10-14 & 0.042 & free \\
\texttt{microsoft/microsoft-decision-1.1} & 2026-11-17 & 0.042 & free \\

\texttt{openai/gpt-6-luna-decisions} & 2026-10-06 & 0.100 & free \\
\texttt{perplexity/pplx-decider-v1.3:free} & 2026-11-12 & 0 & free \\
\texttt{liquid/d1.2-ultra} & 2026-11-16 & 0.042 & free \\
\texttt{cloudflare/clef} & 2026-10-01 & 0.192 & free \\
\texttt{cloudflare/clef-flash} & 2026-10-01 & 0.019 & free \\
\texttt{baseten/mammoth-1} & 2026-10-21 & 0.100 & free \\
\texttt{baseten/elephant-1} & 2026-10-21 & 0.035 & free \\
\texttt{nace-ai/drex-max} & 2026-10-21 & 0.060 & free \\
\texttt{upstage/solar-decide} & 2026-09-28 & 0.050 & free \\
\texttt{scaledown/classify-1} & 2026-10-13 & 0.040 & free \\
\texttt{morph/reflex} & 2026-11-03 & 0.042 & free \\
\texttt{heygen/heygen-decision-1} & 2026-11-13 & 0.140 & free \\
\texttt{respan/span-02} & 2026-10-16 & 0.010 & free \\
\texttt{jaredpalmer/kev-4b} & 2026-09-25 & 0.042 & free \\
\addlinespace
\multicolumn{4}{@{}l}{\textit{Prompted LLMs}} \\
\texttt{anthropic/claude-sonnet-5.5} & -- & 2.00 & 10.00 \\
\texttt{anthropic/claude-haiku-5.5} & -- & 0.10 & 0.50 \\
\texttt{openai/gpt-6.2-sol} & -- & 2.00 & 10.00 \\
\texttt{openai/gpt-6-luna} & -- & 0.10 & 0.50 \\
\texttt{qwen/qwen3.8-27b} & -- & 0.06 & 2.20 \\
\texttt{qwen/qwen3.5-27b} & -- & 0.20 & 1.56 \\
\texttt{qwen/qwen3.5-9b} & -- & 0.15 & 0.23 \\
\bottomrule
\end{tabular}
\tabnote{TDMs bill input tokens only. Perplexity Decider V1.3 is listed only as a free variant, so
its cost per decision is zero.}
\end{threeparttable}
\end{table}

\clearpage
\section{Reliability Diagrams}
\label{app:reliability}

\begin{figure}[H]
\caption{Reliability diagrams for selected systems}
\label{fig:reliability}
\centering
\includegraphics[width=\textwidth]{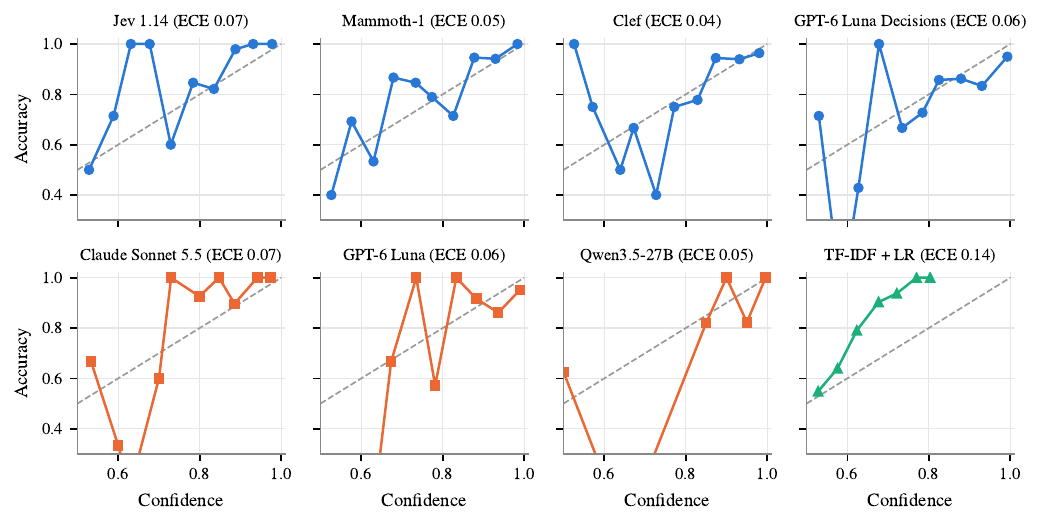}
\tabnote{Ten bins of equal width on the confidence $\max(p, 1-p)$. Bins without rows are omitted.}
\end{figure}

Figure~\ref{fig:reliability} shows reliability diagrams for eight systems. A calibrated system lies
on the diagonal; points below it mean that the system is more confident than it is accurate.

%% file: references.bib
@article{neslin-2006-defection,
  title     = {Defection Detection: {Measuring} and Understanding the Predictive Accuracy of Customer Churn Models},
  author    = {Neslin, Scott A. and Gupta, Sunil and Kamakura, Wagner and Lu, Junxiang and Mason, Charlotte H.},
  journal   = {Journal of Marketing Research},
  volume    = {43},
  number    = {2},
  pages     = {204--211},
  year      = {2006},
  doi       = {10.1509/jmkr.43.2.204},
  url       = {https://doi.org/10.1509/jmkr.43.2.204}
}

@article{verbraken-2013-novel,
  title     = {A Novel Profit Maximizing Metric for Measuring Classification Performance of Customer Churn Prediction Models},
  author    = {Verbraken, Thomas and Verbeke, Wouter and Baesens, Bart},
  journal   = {{IEEE} Transactions on Knowledge and Data Engineering},
  volume    = {25},
  number    = {5},
  pages     = {961--973},
  year      = {2013},
  doi       = {10.1109/TKDE.2012.50},
  url       = {https://doi.org/10.1109/TKDE.2012.50}
}

@misc{manzoor-2025-eprofits,
  title         = {{e-Profits}: {A} Business-Aligned Evaluation Metric for Profit-Sensitive Customer Churn Prediction},
  author        = {Manzoor, Awais and Qureshi, M. Atif and Kidney, Etain and Longo, Luca},
  year          = {2025},
  eprint        = {2507.08860},
  archivePrefix = {arXiv},
  primaryClass  = {cs.LG},
  doi           = {10.48550/arXiv.2507.08860},
  url           = {https://arxiv.org/abs/2507.08860},
  note          = {arXiv preprint, v2 revised 14 January 2026}
}

@article{decaigny-2020-incorporating,
  title     = {Incorporating Textual Information in Customer Churn Prediction Models Based on a Convolutional Neural Network},
  author    = {De Caigny, Arno and Coussement, Kristof and De Bock, Koen W. and Lessmann, Stefan},
  journal   = {International Journal of Forecasting},
  volume    = {36},
  number    = {4},
  pages     = {1563--1578},
  year      = {2020},
  doi       = {10.1016/j.ijforecast.2019.03.029},
  url       = {https://doi.org/10.1016/j.ijforecast.2019.03.029}
}

@article{vo-2021-leveraging,
  title     = {Leveraging Unstructured Call Log Data for Customer Churn Prediction},
  author    = {Vo, Nhi N. Y. and Liu, Shaowu and Li, Xitong and Xu, Guandong},
  journal   = {Knowledge-Based Systems},
  volume    = {212},
  pages     = {106586},
  year      = {2021},
  doi       = {10.1016/j.knosys.2020.106586},
  url       = {https://doi.org/10.1016/j.knosys.2020.106586}
}

@article{shahabikargar-2025-churnkb,
  title     = {{ChurnKB}: {A} Generative {AI}-Enriched Knowledge Base for Customer Churn Feature Engineering},
  author    = {Shahabikargar, Maryam and Beheshti, Amin and Mansoor, Wathiq and Zhang, Xuyun and Foo, Eu Jin and Jolfaei, Alireza and Hanif, Ambreen and Shabani, Nasrin},
  journal   = {Algorithms},
  volume    = {18},
  number    = {4},
  pages     = {238},
  year      = {2025},
  doi       = {10.3390/a18040238},
  url       = {https://doi.org/10.3390/a18040238}
}

@inproceedings{sauter-2026-actionable,
  title     = {Actionable Interpretability for Churn Classification: {A} Text Bottleneck Model Case Study at a Major Telecom Provider},
  author    = {Sauter, Adrian and Neplenbroek, Vera and Vlassopoulos, Georgios and Bardelloni, Gianluigi},
  booktitle = {{Proceedings of the 64th Annual Meeting of the {A}ssociation for {C}omputational {L}inguistics (Volume 6: Industry Track)}},
  pages     = {1005--1024},
  year      = {2026},
  month     = jul,
  address   = {San Diego, California, USA},
  publisher = {Association for Computational Linguistics},
  doi       = {10.18653/v1/2026.acl-industry.70},
  url       = {https://aclanthology.org/2026.acl-industry.70/}
}

@inproceedings{amiri-2015-target,
  title     = {Target-Dependent Churn Classification in Microblogs},
  author    = {Amiri, Hadi and {Daum{\'e} III}, Hal},
  booktitle = {{Proceedings of the Twenty-Ninth {AAAI} Conference on Artificial Intelligence ({AAAI}-15)}},
  volume    = {29},
  number    = {1},
  pages     = {2361--2367},
  year      = {2015},
  publisher = {{AAAI} Press},
  doi       = {10.1609/aaai.v29i1.9532},
  url       = {https://ojs.aaai.org/index.php/AAAI/article/view/9532}
}

@inproceedings{amiri-2016-short,
  title     = {Short Text Representation for Detecting Churn in Microblogs},
  author    = {Amiri, Hadi and {Daum{\'e} III}, Hal},
  booktitle = {{Proceedings of the Thirtieth {AAAI} Conference on Artificial Intelligence ({AAAI}-16)}},
  volume    = {30},
  number    = {1},
  pages     = {2566--2572},
  year      = {2016},
  publisher = {{AAAI} Press},
  doi       = {10.1609/aaai.v30i1.10333},
  url       = {https://ojs.aaai.org/index.php/AAAI/article/view/10333}
}

@inproceedings{abbet-2018-churn,
  title         = {Churn Intent Detection in Multilingual Chatbot Conversations and Social Media},
  author        = {Abbet, Christian and M'hamdi, Meryem and Giannakopoulos, Athanasios and West, Robert and Hossmann, Andreea and Baeriswyl, Michael and Musat, Claudiu},
  booktitle     = {{Proceedings of the 22nd Conference on Computational Natural Language Learning ({CoNLL})}},
  pages         = {161--170},
  year          = {2018},
  month         = oct,
  address       = {Brussels, Belgium},
  publisher     = {Association for Computational Linguistics},
  doi           = {10.18653/v1/K18-1016},
  url           = {https://aclanthology.org/K18-1016/}
}

@inproceedings{jiang-2011-target,
  title     = {Target-dependent {T}witter Sentiment Classification},
  author    = {Jiang, Long and Yu, Mo and Zhou, Ming and Liu, Xiaohua and Zhao, Tiejun},
  booktitle = {{Proceedings of the 49th Annual Meeting of the {A}ssociation for {C}omputational {L}inguistics: Human Language Technologies}},
  pages     = {151--160},
  year      = {2011},
  month     = jun,
  address   = {Portland, Oregon, USA},
  publisher = {Association for Computational Linguistics},
  url       = {https://aclanthology.org/P11-1016/}
}

@inproceedings{pontiki-2014-semeval,
  title     = {{S}em{E}val-2014 Task 4: {Aspect} Based Sentiment Analysis},
  author    = {Pontiki, Maria and Galanis, Dimitris and Pavlopoulos, John and Papageorgiou, Harris and Androutsopoulos, Ion and Manandhar, Suresh},
  booktitle = {{Proceedings of the 8th International Workshop on Semantic Evaluation ({S}em{E}val 2014)}},
  pages     = {27--35},
  year      = {2014},
  month     = aug,
  address   = {Dublin, Ireland},
  publisher = {Association for Computational Linguistics},
  doi       = {10.3115/v1/S14-2004},
  url       = {https://aclanthology.org/S14-2004/}
}

@inproceedings{mohammad-2016-semeval,
  title     = {{S}em{E}val-2016 Task 6: {Detecting} Stance in Tweets},
  author    = {Mohammad, Saif and Kiritchenko, Svetlana and Sobhani, Parinaz and Zhu, Xiaodan and Cherry, Colin},
  booktitle = {{Proceedings of the 10th International Workshop on Semantic Evaluation ({S}em{E}val-2016)}},
  pages     = {31--41},
  year      = {2016},
  month     = jun,
  address   = {San Diego, California},
  publisher = {Association for Computational Linguistics},
  doi       = {10.18653/v1/S16-1003},
  url       = {https://aclanthology.org/S16-1003/}
}

@inproceedings{zhang-2024-sentiment,
  title         = {Sentiment Analysis in the Era of Large Language Models: {A} Reality Check},
  author        = {Zhang, Wenxuan and Deng, Yue and Liu, Bing and Pan, Sinno and Bing, Lidong},
  booktitle     = {{Findings of the {A}ssociation for {C}omputational {L}inguistics: {NAACL} 2024}},
  pages         = {3881--3906},
  year          = {2024},
  month         = jun,
  address       = {Mexico City, Mexico},
  publisher     = {Association for Computational Linguistics},
  doi           = {10.18653/v1/2024.findings-naacl.246},
  url           = {https://aclanthology.org/2024.findings-naacl.246/}
}

@inproceedings{hadifar-2021-million,
  title         = {A Million Tweets Are Worth a Few Points: {Tuning} Transformers for Customer Service Tasks},
  author        = {Hadifar, Amir and Labat, Sofie and Hoste, Veronique and Develder, Chris and Demeester, Thomas},
  booktitle     = {{Proceedings of the 2021 Conference of the {N}orth {A}merican Chapter of the {A}ssociation for {C}omputational {L}inguistics: Human Language Technologies}},
  pages         = {220--225},
  year          = {2021},
  month         = jun,
  address       = {Online},
  publisher     = {Association for Computational Linguistics},
  doi           = {10.18653/v1/2021.naacl-main.21},
  url           = {https://aclanthology.org/2021.naacl-main.21/}
}

@inproceedings{arora-2024-intent,
  title         = {Intent Detection in the Age of {LLM}s},
  author        = {Arora, Gaurav and Jain, Shreya and Merugu, Srujana},
  booktitle     = {{Proceedings of the 2024 Conference on Empirical Methods in Natural Language Processing: Industry Track}},
  pages         = {1559--1570},
  year          = {2024},
  month         = nov,
  address       = {Miami, Florida, US},
  publisher     = {Association for Computational Linguistics},
  doi           = {10.18653/v1/2024.emnlp-industry.114},
  url           = {https://aclanthology.org/2024.emnlp-industry.114/}
}

@inproceedings{brown-2020-language,
  title         = {Language Models are Few-Shot Learners},
  author        = {Brown, Tom and Mann, Benjamin and Ryder, Nick and Subbiah, Melanie and Kaplan, Jared D. and Dhariwal, Prafulla and Neelakantan, Arvind and Shyam, Pranav and Sastry, Girish and Askell, Amanda and Agarwal, Sandhini and Herbert-Voss, Ariel and Krueger, Gretchen and Henighan, Tom and Child, Rewon and Ramesh, Aditya and Ziegler, Daniel and Wu, Jeffrey and Winter, Clemens and Hesse, Chris and Chen, Mark and Sigler, Eric and Litwin, Mateusz and Gray, Scott and Chess, Benjamin and Clark, Jack and Berner, Christopher and McCandlish, Sam and Radford, Alec and Sutskever, Ilya and Amodei, Dario},
  booktitle     = {{Advances in Neural Information Processing Systems}},
  volume        = {33},
  pages         = {1877--1901},
  publisher     = {Curran Associates, Inc.},
  year          = {2020},
  url           = {https://proceedings.neurips.cc/paper/2020/hash/1457c0d6bfcb4967418bfb8ac142f64a-Abstract.html}
}

@article{gilardi-2023-chatgpt,
  title         = {{ChatGPT} Outperforms Crowd Workers for Text-Annotation Tasks},
  author        = {Gilardi, Fabrizio and Alizadeh, Meysam and Kubli, Ma{\"e}l},
  journal       = {Proceedings of the National Academy of Sciences},
  volume        = {120},
  number        = {30},
  pages         = {e2305016120},
  year          = {2023},
  doi           = {10.1073/pnas.2305016120},
  url           = {https://doi.org/10.1073/pnas.2305016120}
}

@article{ziems-2024-can,
  title         = {Can Large Language Models Transform Computational Social Science?},
  author        = {Ziems, Caleb and Held, William and Shaikh, Omar and Chen, Jiaao and Zhang, Zhehao and Yang, Diyi},
  journal       = {Computational Linguistics},
  volume        = {50},
  number        = {1},
  pages         = {237--291},
  year          = {2024},
  doi           = {10.1162/coli_a_00502},
  url           = {https://aclanthology.org/2024.cl-1.8/}
}

@misc{bucher-2024-finetuned,
  title         = {Fine-Tuned `Small' {LLMs} (Still) Significantly Outperform Zero-Shot Generative {AI} Models in Text Classification},
  author        = {Bucher, Martin Juan Jos{\'e} and Martini, Marco},
  year          = {2024},
  eprint        = {2406.08660},
  archivePrefix = {arXiv},
  doi           = {10.48550/arXiv.2406.08660},
  url           = {https://arxiv.org/abs/2406.08660},
  note          = {arXiv preprint, version 2}
}

@inproceedings{edwards-2024-language,
  title         = {Language Models for Text Classification: {Is} In-Context Learning Enough?},
  author        = {Edwards, Aleksandra and Camacho-Collados, Jose},
  booktitle     = {{Proceedings of the 2024 Joint International Conference on Computational Linguistics, Language Resources and Evaluation ({LREC-COLING} 2024)}},
  pages         = {10058--10072},
  year          = {2024},
  publisher     = {ELRA and ICCL},
  url           = {https://aclanthology.org/2024.lrec-main.879/}
}

@inproceedings{pecher-2025-comparing,
  title         = {Comparing Specialised Small and General Large Language Models on Text Classification: 100 Labelled Samples to Achieve Break-Even Performance},
  author        = {Pecher, Branislav and Srba, Ivan and Bielikova, Maria},
  booktitle     = {{Proceedings of the 2025 Conference on Empirical Methods in Natural Language Processing}},
  pages         = {165--184},
  year          = {2025},
  address       = {Suzhou, China},
  publisher     = {Association for Computational Linguistics},
  doi           = {10.18653/v1/2025.emnlp-main.9},
  url           = {https://aclanthology.org/2025.emnlp-main.9/}
}

@article{zschache-2026-comparing,
  title         = {Comparing Energy Consumption and Accuracy in Text Classification Inference},
  author        = {Zschache, Johannes and Hartwig, Tilman},
  journal       = {Scientific Reports},
  volume        = {16},
  pages         = {12717},
  year          = {2026},
  doi           = {10.1038/s41598-026-45023-0},
  url           = {https://doi.org/10.1038/s41598-026-45023-0}
}

@inproceedings{tian-2023-just,
  title         = {Just Ask for Calibration: {Strategies} for Eliciting Calibrated Confidence Scores from Language Models Fine-Tuned with Human Feedback},
  author        = {Tian, Katherine and Mitchell, Eric and Zhou, Allan and Sharma, Archit and Rafailov, Rafael and Yao, Huaxiu and Finn, Chelsea and Manning, Christopher},
  booktitle     = {{Proceedings of the 2023 Conference on Empirical Methods in Natural Language Processing}},
  pages         = {5433--5442},
  year          = {2023},
  address       = {Singapore},
  publisher     = {Association for Computational Linguistics},
  doi           = {10.18653/v1/2023.emnlp-main.330},
  url           = {https://aclanthology.org/2023.emnlp-main.330/}
}

@inproceedings{xiong-2024-can,
  title         = {Can {LLMs} Express Their Uncertainty? {An} Empirical Evaluation of Confidence Elicitation in {LLMs}},
  author        = {Xiong, Miao and Hu, Zhiyuan and Lu, Xinyang and Li, Yifei and Fu, Jie and He, Junxian and Hooi, Bryan},
  booktitle     = {{The Twelfth International Conference on Learning Representations ({ICLR} 2024)}},
  year          = {2024},
  url           = {https://openreview.net/forum?id=gjeQKFxFpZ}
}

@article{salton-1988-term,
  title         = {Term-Weighting Approaches in Automatic Text Retrieval},
  author        = {Salton, Gerard and Buckley, Christopher},
  journal       = {Information Processing \& Management},
  volume        = {24},
  number        = {5},
  pages         = {513--523},
  year          = {1988},
  doi           = {10.1016/0306-4573(88)90021-0},
  url           = {https://doi.org/10.1016/0306-4573(88)90021-0}
}

@inproceedings{reimers-2019-sentencebert,
  title         = {{Sentence-BERT}: {Sentence} Embeddings using {Siamese} {BERT}-Networks},
  author        = {Reimers, Nils and Gurevych, Iryna},
  booktitle     = {{Proceedings of the 2019 Conference on Empirical Methods in Natural Language Processing and the 9th International Joint Conference on Natural Language Processing ({EMNLP-IJCNLP})}},
  pages         = {3982--3992},
  year          = {2019},
  address       = {Hong Kong, China},
  publisher     = {Association for Computational Linguistics},
  doi           = {10.18653/v1/D19-1410},
  url           = {https://aclanthology.org/D19-1410/}
}

@inproceedings{wang-2020-minilm,
  title         = {{MiniLM}: {Deep} Self-Attention Distillation for Task-Agnostic Compression of Pre-Trained Transformers},
  author        = {Wang, Wenhui and Wei, Furu and Dong, Li and Bao, Hangbo and Yang, Nan and Zhou, Ming},
  booktitle     = {{Advances in Neural Information Processing Systems}},
  volume        = {33},
  pages         = {5776--5788},
  publisher     = {Curran Associates, Inc.},
  year          = {2020},
  url           = {https://proceedings.neurips.cc/paper/2020/hash/3f5ee243547dee91fbd053c1c4a845aa-Abstract.html}
}

@article{pedregosa-2011-scikitlearn,
  title         = {Scikit-learn: {Machine} Learning in {Python}},
  author        = {Pedregosa, Fabian and Varoquaux, Ga{\"e}l and Gramfort, Alexandre and Michel, Vincent and Thirion, Bertrand and Grisel, Olivier and Blondel, Mathieu and Prettenhofer, Peter and Weiss, Ron and Dubourg, Vincent and Vanderplas, Jake and Passos, Alexandre and Cournapeau, David and Brucher, Matthieu and Perrot, Matthieu and Duchesnay, {\'E}douard},
  journal       = {Journal of Machine Learning Research},
  volume        = {12},
  number        = {85},
  pages         = {2825--2830},
  year          = {2011},
  url           = {https://jmlr.org/papers/v12/pedregosa11a.html}
}

@article{chen-2024-frugalgpt,
  title         = {{FrugalGPT}: {How} to Use Large Language Models While Reducing Cost and Improving Performance},
  author        = {Chen, Lingjiao and Zaharia, Matei and Zou, James},
  journal       = {Transactions on Machine Learning Research},
  year          = {2024},
  url           = {https://openreview.net/forum?id=cSimKw5p6R}
}

@inproceedings{ong-2025-routellm,
  title         = {{RouteLLM}: {Learning} to Route {LLMs} from Preference Data},
  author        = {Ong, Isaac and Almahairi, Amjad and Wu, Vincent and Chiang, Wei-Lin and Wu, Tianhao and Gonzalez, Joseph E. and Kadous, M. Waleed and Stoica, Ion},
  booktitle     = {{The Thirteenth International Conference on Learning Representations ({ICLR} 2025)}},
  year          = {2025},
  url           = {https://proceedings.iclr.cc/paper_files/paper/2025/hash/5503a7c69d48a2f86fc00b3dc09de686-Abstract-Conference.html}
}

@inproceedings{chen-2025-do,
  title         = {Do {NOT} Think That Much for 2+3=? {On} the Overthinking of Long Reasoning Models},
  author        = {Chen, Xingyu and Xu, Jiahao and Liang, Tian and He, Zhiwei and Pang, Jianhui and Yu, Dian and Song, Linfeng and Liu, Qiuzhi and Zhou, Mengfei and Zhang, Zhuosheng and Wang, Rui and Tu, Zhaopeng and Mi, Haitao and Yu, Dong},
  booktitle     = {{Proceedings of the 42nd International Conference on Machine Learning}},
  series        = {Proceedings of Machine Learning Research},
  volume        = {267},
  pages         = {9487--9499},
  publisher     = {PMLR},
  year          = {2025},
  url           = {https://proceedings.mlr.press/v267/chen25bx.html}
}

@inproceedings{irugalbandara-2024-scaling,
  title         = {Scaling Down to Scale Up: {A} Cost-Benefit Analysis of Replacing {OpenAI}'s {LLM} with Open Source {SLMs} in Production},
  author        = {Irugalbandara, Chandra and Mahendra, Ashish and Daynauth, Roland and Kasthuri Arachchige, Tharuka and Dantanarayana, Jayanaka and Flautner, Krisztian and Tang, Lingjia and Kang, Yiping and Mars, Jason},
  booktitle     = {{2024 {IEEE} International Symposium on Performance Analysis of Systems and Software ({ISPASS})}},
  pages         = {280--291},
  publisher     = {IEEE},
  year          = {2024},
  doi           = {10.1109/ISPASS61541.2024.00034},
  url           = {https://doi.org/10.1109/ISPASS61541.2024.00034}
}

@book{kahneman-2011-thinking,
  author    = {Kahneman, Daniel},
  title     = {Thinking, Fast and Slow},
  publisher = {Farrar, Straus and Giroux},
  address   = {New York},
  year      = {2011},
  isbn      = {9780374275631},
  url       = {https://openlibrary.org/isbn/9780374275631},
  note      = {Verified via Open Library ISBN record; publisher page returned HTTP 403 on 2026-11-19}
}

@article{evans-2013-dualprocess,
  author  = {Evans, Jonathan St. B. T. and Stanovich, Keith E.},
  title   = {Dual-Process Theories of Higher Cognition: {Advancing} the Debate},
  journal = {Perspectives on Psychological Science},
  volume  = {8},
  number  = {3},
  pages   = {223--241},
  year    = {2013},
  doi     = {10.1177/1745691612460685},
  url     = {https://journals.sagepub.com/doi/10.1177/1745691612460685}
}

@article{guo-2025-deepseekr1,
  author        = {Guo, Daya and Yang, Dejian and Zhang, Haowei and Song, Junxiao and Wang, Peiyi and Zhu, Qihao and Xu, Runxin and Zhang, Ruoyu and Ma, Shirong and Bi, Xiao and Zhang, Xiaokang and Yu, Xingkai and Wu, Yu and Wu, Z. F. and Gou, Zhibin and Shao, Zhihong and Li, Zhuoshu and Gao, Ziyi and Liu, Aixin and others},
  title         = {{DeepSeek-R1} Incentivizes Reasoning in {LLMs} through Reinforcement Learning},
  journal       = {Nature},
  volume        = {645},
  number        = {8081},
  pages         = {633--638},
  year          = {2025},
  doi           = {10.1038/s41586-025-09422-z},
  url           = {https://doi.org/10.1038/s41586-025-09422-z}
}

@misc{li-2025-from,
  author        = {Li, Zhong-Zhi and Zhang, Duzhen and Zhang, Ming-Liang and Zhang, Jiaxin and Liu, Zengyan and Yao, Yuxuan and Xu, Haotian and Zheng, Junhao and Wang, Pei-Jie and Chen, Xiuyi and Zhang, Yingying and Yin, Fei and Dong, Jiahua and Li, Zhiwei and Bi, Bao-Long and Mei, Ling-Rui and Fang, Junfeng and Liang, Xiao and Guo, Zhijiang and Song, Le and Liu, Cheng-Lin},
  title         = {From {System} 1 to {System} 2: {A} Survey of Reasoning Large Language Models},
  year          = {2025},
  eprint        = {2502.17419},
  archivePrefix = {arXiv},
  primaryClass  = {cs.AI},
  doi           = {10.48550/arXiv.2502.17419},
  url           = {https://arxiv.org/abs/2502.17419},
  note          = {Version 6}
}

@misc{almeida-2026-introducing,
  author       = {Almeida, Diogo},
  title        = {Introducing {System One} Models \& {Jev}},
  howpublished = {TypeSafe AI Blog},
  year         = {2026},
  month        = sep,
  day          = {15},
  url          = {https://typesafe.ai/blog/introducing-system-one-models-and-jev},
  urldate      = {2026-11-19}
}

@misc{typesafe-2026-system,
  author       = {{TypeSafe AI}},
  title        = {{System One}},
  howpublished = {TypeSafe documentation},
  year         = {2026},
  url          = {https://docs.typesafe.ai/concepts/system-one},
  urldate      = {2026-11-19}
}

@misc{typesafe-2026-confidence,
  author       = {{TypeSafe AI}},
  title        = {Confidence},
  howpublished = {TypeSafe documentation},
  year         = {2026},
  url          = {https://docs.typesafe.ai/confidence},
  urldate      = {2026-11-19}
}

@misc{typesafe-2026-model,
  author       = {{TypeSafe AI}},
  title        = {Model Jaggedness},
  howpublished = {TypeSafe documentation},
  year         = {2026},
  url          = {https://docs.typesafe.ai/model-jaggedness},
  urldate      = {2026-11-19}
}

@misc{openrouter-2026-decisions,
  author       = {{OpenRouter}},
  title        = {Decisions {API}},
  howpublished = {OpenRouter documentation (alpha)},
  year         = {2026},
  url          = {https://openrouter.ai/docs/guides/features/decisions},
  urldate      = {2026-11-19}
}

@misc{openrouter-2026-models,
  author       = {{OpenRouter}},
  title        = {Models {API}: {Models} with Output Modality ``decisions''},
  howpublished = {API listing (JSON), 44 entries on access date},
  year         = {2026},
  url          = {https://openrouter.ai/api/v1/models?output_modalities=decisions},
  urldate      = {2026-11-19}
}

@misc{chen-2026-introducing,
  author       = {Chen, Michelle and Reneau, Alex and Flansburg, Kevin},
  title        = {Introducing {Clef}: {Our} Open-Source Decision Models, and New {RL} Fine-Tuning Platform},
  howpublished = {The Cloudflare Blog},
  year         = {2026},
  month        = oct,
  day          = {1},
  url          = {https://blog.cloudflare.com/clef-decision-models/},
  urldate      = {2026-11-19}
}

@misc{baselabs-2026-mammoth1,
  author       = {{Base Labs at Baseten}},
  title        = {mammoth-1},
  howpublished = {Hugging Face model card, revision 61cec44c281c693992e3d6ebeb44dd3402dfa186},
  year         = {2026},
  url          = {https://huggingface.co/baselabs/mammoth-1},
  urldate      = {2026-11-19},
  note         = {License MIT; base model Qwen/Qwen3.5-27B}
}

@misc{rafe-2026-benchmarking,
  author        = {Rafe, Amir and Das, Subasish},
  title         = {Benchmarking {System One} Decision Models against Trained Classifiers and Language Models for Automated Decision Gates},
  year          = {2026},
  eprint        = {2610.00346},
  archivePrefix = {arXiv},
  primaryClass  = {cs.LG},
  doi           = {10.48550/arXiv.2610.00346},
  url           = {https://arxiv.org/abs/2610.00346},
  note          = {Version 2, 6 Oct 2026}
}

@misc{chen-2026-system1arena,
  author        = {Chen, Tianyu and Liu, Xinyang and An, Ruyi and Wang, Zhendong and Zhou, Mingyuan},
  title         = {{System1-Arena}: {Which} Decisions Can Be Left to {System} 1?},
  year          = {2026},
  eprint        = {2610.21244},
  archivePrefix = {arXiv},
  primaryClass  = {cs.CL},
  doi           = {10.48550/arXiv.2610.21244},
  url           = {https://arxiv.org/abs/2610.21244}
}

@misc{duan-2026-general,
  author        = {Duan, Feiyu and Lin, Jiayu and Wang, Jia and Xiang, Jun and Wu, Jialiang and Zhang, Xinnong and Yan, Hanqi and Wang, Siyuan and Wei, Zhongyu},
  title         = {General Decision Models: {Benchmarking} and Insights Beyond {Jev}},
  year          = {2026},
  eprint        = {2610.03935},
  archivePrefix = {arXiv},
  primaryClass  = {cs.CL},
  doi           = {10.48550/arXiv.2610.03935},
  url           = {https://arxiv.org/abs/2610.03935},
  note          = {Version 2}
}

@misc{ibrahim-2026-evaluating,
  author        = {Ibrahim, Hazem and Zaki, Yasir},
  title         = {Evaluating Decision Models for Text Annotation in Computational Social Science},
  year          = {2026},
  eprint        = {2609.24574},
  archivePrefix = {arXiv},
  primaryClass  = {cs.CL},
  doi           = {10.48550/arXiv.2609.24574},
  url           = {https://arxiv.org/abs/2609.24574},
  note          = {Version 2}
}

@misc{roumeliotis-2026-decision,
  author       = {Roumeliotis, Konstantinos I. and Margaris, Dionisis and Spiliotopoulos, Dimitris and Tselikas, Nikolaos D.},
  title        = {Decision Models for Business Customer-Service Triage: {Accuracy}, Calibration, Out-of-Scope Detection and Cost of {Jev}, {Clef} and {Perplexity Decider}},
  howpublished = {Preprints.org},
  year         = {2026},
  month        = oct,
  doi          = {10.20944/preprints202610.1480.v1},
  url          = {https://www.preprints.org/manuscript/202610.1480/v1},
  note         = {Preprint, posted 2026-10-15. Verified via Crossref DOI record; landing page returned HTTP 403 on 2026-11-19}
}

@misc{suppa-2026-do,
  author        = {{\v{S}}uppa, Marek and Kop{\v{c}}an, Jaroslav},
  title         = {Do We Still Need Fine-Tuned Classifiers? {Swapping} Typed Decision Models into 24 {ACL}-Venue Papers},
  year          = {2026},
  eprint        = {2611.20191},
  archivePrefix = {arXiv},
  primaryClass  = {cs.CL},
  doi           = {10.48550/arXiv.2611.20191},
  url           = {https://arxiv.org/abs/2611.20191}
}

@misc{li-2026-can,
  author        = {Li, Xing and Ning, Jinzhong and Zhang, Yijia and Yang, Liang and Lin, Hongfei},
  title         = {Can Decision Models Understand Stance? {Evaluating} {Jev} Against General-Purpose {LLMs}},
  year          = {2026},
  eprint        = {2610.11901},
  archivePrefix = {arXiv},
  primaryClass  = {cs.CL},
  doi           = {10.48550/arXiv.2610.11901},
  url           = {https://arxiv.org/abs/2610.11901}
}

@misc{brom-2026-confidence,
  author        = {Brom, Pedro Carvalho},
  title         = {Confidence Is Not a Promise: {A} Reference Audit of Typed Decision Models},
  year          = {2026},
  eprint        = {2610.36311},
  archivePrefix = {arXiv},
  primaryClass  = {cs.LG},
  doi           = {10.48550/arXiv.2610.36311},
  url           = {https://arxiv.org/abs/2610.36311}
}

@misc{sun-2026-typesafe,
  author        = {Sun, Yu and Xu, Junhao and Shi, Jiajia and Yang, Zijin},
  title         = {Type-Safe Is Not Error-Free: {Typed} Decision Models Follow the Option Name, Not the Definition Bound to It},
  year          = {2026},
  eprint        = {2609.26758},
  archivePrefix = {arXiv},
  primaryClass  = {cs.AI},
  doi           = {10.48550/arXiv.2609.26758},
  url           = {https://arxiv.org/abs/2609.26758},
  note          = {Version 3}
}

@article{brier-1950-verification,
  author    = {Brier, Glenn W.},
  title     = {Verification of Forecasts Expressed in Terms of Probability},
  journal   = {Monthly Weather Review},
  volume    = {78},
  number    = {1},
  pages     = {1--3},
  year      = {1950},
  doi       = {10.1175/1520-0493(1950)078<0001:VOFEIT>2.0.CO;2},
  url       = {https://doi.org/10.1175/1520-0493(1950)078<0001:VOFEIT>2.0.CO;2}
}

@article{degroot-1983-comparison,
  author    = {DeGroot, Morris H. and Fienberg, Stephen E.},
  title     = {The Comparison and Evaluation of Forecasters},
  journal   = {Journal of the Royal Statistical Society: Series D (The Statistician)},
  volume    = {32},
  number    = {1/2},
  pages     = {12--22},
  year      = {1983},
  doi       = {10.2307/2987588},
  url       = {https://doi.org/10.2307/2987588}
}

@inproceedings{niculescumizil-2005-predicting,
  author    = {Niculescu-Mizil, Alexandru and Caruana, Rich},
  title     = {Predicting Good Probabilities with Supervised Learning},
  booktitle = {{Proceedings of the 22nd International Conference on Machine Learning ({ICML} '05)}},
  publisher = {ACM},
  address   = {Bonn, Germany},
  pages     = {625--632},
  year      = {2005},
  doi       = {10.1145/1102351.1102430},
  url       = {https://doi.org/10.1145/1102351.1102430}
}

@inproceedings{naeini-2015-obtaining,
  author    = {Pakdaman Naeini, Mahdi and Cooper, Gregory F. and Hauskrecht, Milos},
  title     = {Obtaining Well Calibrated Probabilities Using {B}ayesian Binning},
  booktitle = {{Proceedings of the {AAAI} Conference on Artificial Intelligence}},
  volume    = {29},
  number    = {1},
  pages     = {2901--2907},
  year      = {2015},
  doi       = {10.1609/aaai.v29i1.9602},
  url       = {https://doi.org/10.1609/aaai.v29i1.9602}
}

@inproceedings{guo-2017-calibration,
  author    = {Guo, Chuan and Pleiss, Geoff and Sun, Yu and Weinberger, Kilian Q.},
  title     = {On Calibration of Modern Neural Networks},
  booktitle = {{Proceedings of the 34th International Conference on Machine Learning ({ICML})}},
  series    = {Proceedings of Machine Learning Research},
  volume    = {70},
  pages     = {1321--1330},
  publisher = {PMLR},
  year      = {2017},
  url       = {https://proceedings.mlr.press/v70/guo17a.html}
}

@inproceedings{nixon-2019-measuring,
  author    = {Nixon, Jeremy and Dusenberry, Michael W. and Zhang, Linchuan and Jerfel, Ghassen and Tran, Dustin},
  title     = {Measuring Calibration in Deep Learning},
  booktitle = {{Proceedings of the {IEEE/CVF} Conference on Computer Vision and Pattern Recognition ({CVPR}) Workshops}},
  pages     = {38--41},
  year      = {2019},
  url       = {https://openaccess.thecvf.com/content_CVPRW_2019/html/Uncertainty_and_Robustness_in_Deep_Visual_Learning/Nixon_Measuring_Calibration_in_Deep_Learning_CVPRW_2019_paper.html},
  note      = {Extended arXiv version (v2) has additional authors and metrics}
}

@inproceedings{kumar-2019-verified,
  author    = {Kumar, Ananya and Liang, Percy and Ma, Tengyu},
  title     = {Verified Uncertainty Calibration},
  booktitle = {{Advances in Neural Information Processing Systems}},
  volume    = {32},
  publisher = {Curran Associates, Inc.},
  year      = {2019},
  url       = {https://proceedings.neurips.cc/paper_files/paper/2019/hash/f8c0c968632845cd133308b1a494967f-Abstract.html}
}

@article{chow-1970-optimum,
  author    = {Chow, C.},
  title     = {On Optimum Recognition Error and Reject Tradeoff},
  journal   = {IEEE Transactions on Information Theory},
  volume    = {16},
  number    = {1},
  pages     = {41--46},
  year      = {1970},
  doi       = {10.1109/TIT.1970.1054406},
  url       = {https://doi.org/10.1109/TIT.1970.1054406}
}

@article{elyaniv-2010-foundations,
  author    = {El-Yaniv, Ran and Wiener, Yair},
  title     = {On the Foundations of Noise-free Selective Classification},
  journal   = {Journal of Machine Learning Research},
  volume    = {11},
  number    = {53},
  pages     = {1605--1641},
  year      = {2010},
  url       = {https://www.jmlr.org/papers/v11/el-yaniv10a.html}
}

@inproceedings{geifman-2017-selective,
  author    = {Geifman, Yonatan and El-Yaniv, Ran},
  title     = {Selective Classification for Deep Neural Networks},
  booktitle = {{Advances in Neural Information Processing Systems}},
  volume    = {30},
  publisher = {Curran Associates, Inc.},
  year      = {2017},
  url       = {https://proceedings.neurips.cc/paper_files/paper/2017/hash/4a8423d5e91fda00bb7e46540e2b0cf1-Abstract.html}
}

@inproceedings{kamath-2020-selective,
  author    = {Kamath, Amita and Jia, Robin and Liang, Percy},
  title     = {Selective Question Answering under Domain Shift},
  booktitle = {{Proceedings of the 58th Annual Meeting of the {A}ssociation for {C}omputational {L}inguistics}},
  pages     = {5684--5696},
  publisher = {Association for Computational Linguistics},
  address   = {Online},
  year      = {2020},
  doi       = {10.18653/v1/2020.acl-main.503},
  url       = {https://aclanthology.org/2020.acl-main.503/}
}

@article{brynjolfsson-2025-generative,
  author    = {Brynjolfsson, Erik and Li, Danielle and Raymond, Lindsey},
  title     = {Generative {AI} at Work},
  journal   = {The Quarterly Journal of Economics},
  volume    = {140},
  number    = {2},
  pages     = {889--942},
  year      = {2025},
  doi       = {10.1093/qje/qjae044},
  url       = {https://doi.org/10.1093/qje/qjae044}
}

@article{baayen-2008-mixed,
  author    = {Baayen, R. H. and Davidson, D. J. and Bates, D. M.},
  title     = {Mixed-Effects Modeling with Crossed Random Effects for Subjects and Items},
  journal   = {Journal of Memory and Language},
  volume    = {59},
  number    = {4},
  pages     = {390--412},
  year      = {2008},
  doi       = {10.1016/j.jml.2007.12.005},
  url       = {https://doi.org/10.1016/j.jml.2007.12.005}
}

@article{barr-2013-random,
  author    = {Barr, Dale J. and Levy, Roger and Scheepers, Christoph and Tily, Harry J.},
  title     = {Random Effects Structure for Confirmatory Hypothesis Testing: {Keep} It Maximal},
  journal   = {Journal of Memory and Language},
  volume    = {68},
  number    = {3},
  pages     = {255--278},
  year      = {2013},
  doi       = {10.1016/j.jml.2012.11.001},
  url       = {https://doi.org/10.1016/j.jml.2012.11.001}
}

@article{bates-2015-fitting,
  author    = {Bates, Douglas and M{\"a}chler, Martin and Bolker, Ben and Walker, Steve},
  title     = {Fitting Linear Mixed-Effects Models Using {lme4}},
  journal   = {Journal of Statistical Software},
  volume    = {67},
  number    = {1},
  pages     = {1--48},
  year      = {2015},
  doi       = {10.18637/jss.v067.i01},
  url       = {https://www.jstatsoft.org/article/view/v067i01}
}

@manual{lenth-2026-emmeans,
  author    = {Lenth, Russell V. and Piaskowski, Julia},
  title     = {{emmeans}: {Estimated} Marginal Means, aka Least-Squares Means},
  note      = {R package version 2.0.2},
  organization = {Comprehensive R Archive Network (CRAN)},
  year      = {2026},
  doi       = {10.32614/CRAN.package.emmeans},
  url       = {https://CRAN.R-project.org/package=emmeans}
}

@article{burkner-2017-brms,
  author    = {B{\"u}rkner, Paul-Christian},
  title     = {{brms}: {An} {R} Package for {B}ayesian Multilevel Models Using {Stan}},
  journal   = {Journal of Statistical Software},
  volume    = {80},
  number    = {1},
  pages     = {1--28},
  year      = {2017},
  doi       = {10.18637/jss.v080.i01},
  url       = {https://www.jstatsoft.org/article/view/v080i01}
}

@inproceedings{hagmann-2023-towards,
  author    = {Hagmann, Michael and Meier, Philipp and Riezler, Stefan},
  title     = {Towards Inferential Reproducibility of Machine Learning Research},
  booktitle = {{The Eleventh International Conference on Learning Representations ({ICLR})}},
  year      = {2023},
  url       = {https://openreview.net/forum?id=li4GQCQWkv}
}

@article{field-2007-bootstrapping,
  author    = {Field, C. A. and Welsh, A. H.},
  title     = {Bootstrapping Clustered Data},
  journal   = {Journal of the Royal Statistical Society: Series B (Statistical Methodology)},
  volume    = {69},
  number    = {3},
  pages     = {369--390},
  year      = {2007},
  doi       = {10.1111/j.1467-9868.2007.00593.x},
  url       = {https://doi.org/10.1111/j.1467-9868.2007.00593.x}
}
